\documentclass[11pt,a4paper,dvipsnames]{article}
\usepackage[hyperref]{emnlp2018}
\usepackage{times}
\usepackage{latexsym}
\usepackage{url}
\usepackage{color}
\usepackage{graphicx}
\usepackage{enumitem}
\usepackage{pgfplots}
\usepgfplotslibrary{colorbrewer}
\pgfplotsset{compat=1.14}
\pgfplotsset{cycle list/Dark2}

\newcommand{\dataset}[1]{\textsc{#1}}

\aclfinalcopy 

\title{How clever is the FiLM model, and how clever can it be?}

\author{
Alexander Kuhnle \\
Department of Computer \\
Science and Technology \\
University of Cambridge \\
{\tt aok25@cam.ac.uk} \\\And
Huiyuan Xie \\
Department of Computer \\
Science and Technology \\
University of Cambridge \\
{\tt hx255@cam.ac.uk} \\\And
Ann Copestake \\
Department of Computer \\
Science and Technology \\
University of Cambridge \\
{\tt aac10@cam.ac.uk} \\
}
\date{}

\begin{document}

\maketitle

\begin{abstract}
The FiLM model achieves close-to-perfect performance on the diagnostic CLEVR dataset and is distinguished from other such models by having a comparatively simple and easily transferable architecture. In this paper, we investigate in more detail the ability of FiLM to learn various linguistic constructions. Our main results show that (a) FiLM is not able to learn relational statements straight away except for very simple instances, (b) training on a broader set of instances as well as pretraining on simpler instance types can help alleviate these learning difficulties, (c) mixing is less robust than pretraining and very sensitive to the compositional structure of the dataset. Overall, our results suggest that the approach of big all-encompassing datasets and the paradigm of \textit{``the effectiveness of data''} may have fundamental limitations.
\end{abstract}

\section{Introduction}

The task of Visual Question Answering (VQA) lies at the intersection of Computer Vision and Natural Language Processing. It generalizes the vision tasks of object detection/recognition to arbitrary visual-linguistic inferences, limited only by what can be queried by language. At the same time, systems can more easily be evaluated than this is the case for related multimodal tasks like image captioning. The task became popular in recent years, particularly the VQA Dataset \cite{Antol2015}, based on which a third edition of the VQA Challenge is run this year.

In reaction to various issues that allow comparatively naive models -- for instance, a text-only system ignoring visual information and solely relying on language statistics -- to achieve competitive performance on the VQA Dataset \cite{Agrawal2016,Goyal2017,Mudrakarta2018}, abstract and (semi-)automatically generated datasets were introduced \cite{Johnson2017a,Kuhnle2017,Suhr2017,Yang2018}. Their motivation is to provide diagnostic tasks, with the aim of analyzing core abilities for visually grounded language understanding like spatial reasoning or counting, and which cannot be `cheated' as easily by just relying on surface statistics. CLEVR \cite{Johnson2017a} is the most widely adopted of these, and several systems by now achieve near-perfect performance on it \cite{Hu2017,Johnson2017b,Santoro2017,Perez2018,Suarez2018,Hudson2018,Mascharka2018,Yang2018}.

One of the advantages of CLEVR is that it annotates questions from a set of instance types, like \textit{``count''} or \textit{``compare attribute''}, which makes a more detailed evaluation and model comparison possible. Building on the `unit-testing' proposal of \newcite{Kuhnle2018} and related work such as the bAbI tasks for reading comprehension \cite{Weston2015}, which further emphasize the importance and value of targeted evaluation, we analyzed the FiLM model \cite{Perez2018} on the ShapeWorld evaluation framework \cite{Kuhnle2017}. In doing so, we aim to investigate whether its close-to-perfect performance on CLEVR translates to ShapeWorld data as expected, and to shed more light on the strengths and weaknesses of FiLM in general.

\begin{figure*}
\centering\small
\begin{description}[itemsep=0.025cm]
\item[Existential:] ``There is a red square.'', ``A red shape is a square.''
\item[Single-shape:] same as above, with only one object present
\item[Logical:] two existential statements connected by: and, or, if, if and only if 
\item[Numbers:] zero to five; with modifiers: less/more than, at most/least, exactly, not
\item[Quantifiers:] with modifiers as above: no, half, all, a/two third(s), a/three quarter(s)
\item[Relational:] left, right, above, below, 
closer, farther, darker, lighter, smaller, bigger, same/different shape/color
\item[Simple-spatial:] the first four spatial relations, with only two objects per scene
\item[Relational-negation:] relational plus negated relations
\item[Implicit-relational:] left, right, upper, lower, smaller, bigger, darker, lighter, closer, farther (of two target objects)
\item[Superlatives:] superlative forms of the above, of an arbitrary number of target objects
\end{description}
\vspace{0.2cm}
\begin{minipage}[t]{0.38\linewidth}
\begin{center}
\textbf{\normalsize Examples for visual scenes}
\end{center}
\begin{minipage}{0.48\linewidth}
\includegraphics[width=\linewidth]{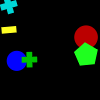}
\end{minipage}%
\hspace{0.04\linewidth}%
\begin{minipage}{0.48\linewidth}
\includegraphics[width=\linewidth]{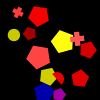}
\end{minipage}
\end{minipage}%
\hspace{0.02\linewidth}%
\begin{minipage}[t]{0.56\linewidth}
\begin{center}
\textbf{\normalsize Examples for true or false statements}
\end{center}
\vspace{-0.3cm}
\begin{itemize}[itemsep=0cm,label=$\circ$]
\item \textit{``There is a cyan square or a circle is green.''}
\item \textit{``At least two shapes are green.''}
\item \textit{``More than half the pentagons are red.''}
\item \textit{``A red cross is to the left of a yellow shape.''}
\item \textit{``The left circle is blue.''}
\item \textit{``The lowermost yellow shape is a circle.''}
\end{itemize}
\end{minipage}
\caption{\textit{Top}: All basic datasets we experimented with, together with their central words/constructions. \textit{Bottom left}: Two example visual scenes. \textit{Bottom right}: Some example captions taken from different datasets (\dataset{logical}, \dataset{numbers}, \dataset{quantifiers}, \dataset{relational}, \dataset{implicit-relational}, \dataset{superlatives}).}
\label{figure:examples}
\end{figure*}

Why FiLM? Arguably, it is one of the simplest models on that performance level for CLEVR. In particular, it does not rely on the semantic program trees underlying its instances, as compared to \cite{Hu2017,Mascharka2018,Johnson2017b}. The first two strictly require the CLEVR-specific program vocabulary, which is different from the one used by ShapeWorld to generate data. The latter is agnostic to the vocabulary, but still sensitive to the size of the vocabulary, which is bigger for ShapeWorld (we ran into memory issues when trying to run this model on ShapeWorld data). Moreover, the code for FiLM is open-source, and in our experiments we found that the model shows robust learning behavior on ShapeWorld data without any tuning of the CLEVR-based default hyperparameters.

While FiLM manages to solve many tasks perfectly, it fails to learn anything on almost all datasets consisting of relational statements. We investigate how two approaches -- broader training sets including simpler instances, and a version of curriculum learning \cite{Elman1993,Bengio2009} -- can make the difference between no learning at all and perfectly solving these datasets. However, we find that the first approach is very sensitive to details of the dataset structure.

These results put into question the common assumption of \textit{``the effectiveness of data''} \cite{Halevy2009} underlying datasets such as the VQA Dataset \cite{Antol2015}, or SQuAD \cite{Rajpurkar2016} for reading comprehension, or SNLI \cite{Bowman2015b} for language inference: that all necessary abilities for a task can simply be learned from one big all-encompassing dataset, and that more data should lead to improved performance. Curriculum learning, on the other hand, shows promise as a robust approach to solving more complex instances of a task.

\begin{figure*}
\newcommand{\good}[1]{\textcolor{green!70!black}{#1}}
\newcommand{\okay}[1]{\textcolor{YellowOrange!80!black}{#1}}
\newcommand{\bad}[1]{\textcolor{red!70!black}{#1}}
\centering
\resizebox{0.46\linewidth}{!}{
\begin{tikzpicture}[baseline=(current bounding box.center)]
\begin{axis}[
every axis plot/.append style={line width=2.5pt},
cycle list name=Dark2,
mark options={mark=none},
width=10cm,
height=7.5cm,
xmin=0, xmax=100,
xtick={0,20,40,60,80,100},
xticklabel style={font=\footnotesize},
xmajorgrids,
ymin=0.49, ymax=1.0,
ytick={0.5,0.6,0.7,0.8,0.9,1.0},
ymajorgrids,
yticklabel style={font=\footnotesize},
grid style=dashed,
legend style={at={(1.0,0.57)},font=\scriptsize,row sep=-2pt}
]
\addplot+[color=CarnationPink] coordinates {
(0,0.5065) (1,0.8265) (2,0.8467) (3,0.9147) (4,0.9446) (5,0.9601) (6,0.9685) (7,0.9862) (8,0.9839) (9,0.9932) (10,0.9976) (15,0.9955) (20,1.0) (25,0.9994) (30,0.9989) (35,0.9993) (40,0.9999) (45,0.9914) (50,1.0) (55,0.9999) (60,0.999) (65,0.9996) (70,0.9998) (75,1.0) (80,1.0) (85,1.0) (90,0.9998) (95,1.0) (100,1.0)
};
\pgfplotsset{cycle list shift=-1}
\addplot coordinates {
(0,0.5008) (1,0.678) (2,0.7294) (3,0.8055) (4,0.9026) (5,0.9443) (6,0.9769) (7,0.9798) (8,0.9828) (9,0.9919) (10,0.9962) (15,0.997) (20,0.9891) (25,0.999) (30,0.9991) (35,0.9954) (40,0.9997) (45,0.6312) (50,0.9998) (55,0.9993) (60,0.9977) (65,0.9984) (70,0.9996) (75,0.9977) (80,0.997) (85,0.9999) (90,0.9995) (95,0.9993) (100,0.9996)
};
\addplot coordinates {
(0,0.5062) (1,0.6154) (2,0.6288) (3,0.6394) (4,0.6427) (5,0.6497) (6,0.6502) (7,0.6576) (8,0.6555) (9,0.6315) (10,0.6554) (15,0.661) (20,0.6722) (25,0.6826) (30,0.6935) (35,0.7213) (40,0.7708) (45,0.8604) (50,0.9377) (55,0.9698) (60,0.9848) (65,0.9887) (70,0.994) (75,0.9958) (80,0.9964) (85,0.9977) (90,0.9972) (95,0.9991) (100,0.9986)
};
\addplot coordinates {
(0,0.5093) (1,0.5342) (2,0.5491) (3,0.5624) (4,0.548) (5,0.5685) (6,0.5657) (7,0.5785) (8,0.5872) (9,0.5686) (10,0.5855) (15,0.6223) (20,0.6983) (25,0.7234) (30,0.7235) (35,0.759) (40,0.861) (45,0.9019) (50,0.9578) (55,0.9664) (60,0.9838) (65,0.9897) (70,0.9909) (75,0.9915) (80,0.9946) (85,0.9879) (90,0.9915) (95,0.9923) (100,0.9959)
};
\addplot coordinates {
(0,0.494) (1,0.5745) (2,0.5724) (3,0.5618) (4,0.5695) (5,0.5779) (6,0.5894) (7,0.5857) (8,0.5941) (9,0.5999) (10,0.6073) (15,0.6577) (20,0.6594) (25,0.6698) (30,0.683) (35,0.7193) (40,0.7393) (45,0.7813) (50,0.8317) (55,0.9319) (60,0.9462) (65,0.9587) (70,0.9544) (75,0.9722) (80,0.9688) (85,0.9774) (90,0.9719) (95,0.9696) (100,0.977)
};
\pgfplotsset{cycle list shift=2}
\addplot coordinates {
(0,0.5056) (1,0.5627) (2,0.5812) (3,0.5622) (4,0.6491) (5,0.6432) (6,0.6472) (7,0.6515) (8,0.6338) (9,0.6494) (10,0.6507) (15,0.6493) (20,0.6599) (25,0.6933) (30,0.7224) (35,0.7263) (40,0.7329) (45,0.7307) (50,0.7347) (55,0.7342) (60,0.7342) (65,0.7374) (70,0.7422) (75,0.7388) (80,0.7526) (85,0.8062) (90,0.8323) (95,0.8416) (100,0.8509)
};
\pgfplotsset{cycle list shift=-2}
\addplot coordinates {
(0,0.494) (1,0.5043) (2,0.5005) (3,0.5147) (4,0.5129) (5,0.5112) (6,0.495) (7,0.5174) (8,0.5028) (9,0.5198) (10,0.4937) (15,0.5115) (20,0.4951) (25,0.494) (30,0.5032) (35,0.4915) (40,0.5056) (45,0.5019) (50,0.5046) (55,0.4955) (60,0.5064) (65,0.5025) (70,0.5054) (75,0.5062) (80,0.5061) (85,0.4947) (90,0.5065) (95,0.509) (100,0.506)
};
\addplot coordinates {
(0,0.4987) (1,0.5211) (2,0.502) (3,0.5108) (4,0.5302) (5,0.53) (6,0.5142) (7,0.5169) (8,0.518) (9,0.529) (10,0.527) (15,0.5162) (20,0.531) (25,0.5234) (30,0.5282) (35,0.5254) (40,0.5037) (45,0.5027) (50,0.5205) (55,0.5179) (60,0.4948) (65,0.527) (70,0.4988) (75,0.5259) (80,0.5304) (85,0.5289) (90,0.5097) (95,0.5285) (100,0.5285)
};
\addplot coordinates {
(0,0.5004) (1,0.4923) (2,0.5014) (3,0.5008) (4,0.5023) (5,0.5017) (6,0.5024) (7,0.517) (8,0.5016) (9,0.5067) (10,0.5006) (15,0.5107) (20,0.511) (25,0.4955) (30,0.5006) (35,0.5224) (40,0.5004) (45,0.5128) (50,0.5092) (55,0.5107) (60,0.5118) (65,0.5076) (70,0.5103) (75,0.5004) (80,0.4993) (85,0.5021) (90,0.5063) (95,0.5033) (100,0.5078)
};
\legend{single-shape,existential,logical,numbers, quantifiers,simple-spatial,relational,implicit-rel,superlatives}
\end{axis}
\end{tikzpicture}}%
\hspace{0.04\linewidth}%
\resizebox{0.46\linewidth}{!}{
\setlength{\tabcolsep}{0.1cm}%
\renewcommand{\arraystretch}{1.25}%
\begin{tabular}[c]{|l|cc|cc|cc|}
\hline
Dataset & \multicolumn{2}{|c|}{CNN-LSTM} & \multicolumn{2}{|c|}{\scalebox{0.75}{CNN-LSTM-SA}} & \multicolumn{2}{|c|}{FiLM} \\
\hline
single-shape
& \multicolumn{2}{|c|}{---} & \multicolumn{2}{|c|}{---} & \good{100.0} & \okay{87.2} \\
existential
& \good{100.0} & \okay{81.1} & \good{100.0} & \good{99.7} & \good{100.0} & \good{99.9} \\
logical
& \okay{79.7} & \bad{62.2} & \okay{76.5} & \bad{58.4} & \good{99.9} & \good{98.9} \\
numbers
& \okay{75.0} & \bad{66.4} & \good{99.1} & \good{98.2} & \good{99.6} & \good{99.3} \\
quantifiers
& \bad{72.1} & \bad{69.1} & \okay{84.8} & \okay{80.8} & \good{97.7} & \good{97.0} \\
simple-spatial
& \okay{81.4} & \bad{64.8} & \okay{81.9} & \bad{57.7} & \okay{85.1} & \bad{61.3} \\
relational
& \multicolumn{2}{|c|}{---} & \multicolumn{2}{|c|}{---} & \bad{50.6} & \bad{51.0} \\
implicit-rel
& \multicolumn{2}{|c|}{---} & \multicolumn{2}{|c|}{---} & \bad{52.9} & \bad{53.2} \\
superlatives
& \multicolumn{2}{|c|}{---} & \multicolumn{2}{|c|}{---} & \bad{50.8} & \bad{50.2} \\
\hline
\end{tabular}}
\caption{
\textit{Left diagram}: validation performance of the FiLM model trained on various ShapeWorld datasets (\textit{x-axis}: iterations in 1000, \textit{y-axis}: accuracy). \textit{Right table}: final validation (\textit{left}) and test (\textit{right}) performance of the trained FiLM models, plus performance of the two baselines CNN-LSTM and CNN-LSTM-SA on selected datasets (accuracy in percent, \textit{\good{green}}: $\geq\! 95\%$, \textit{\okay{orange}}: $\geq\! 75\%$, \textit{\bad{red}}: $<\! 75\%$).}
\label{figure:performance1}
\end{figure*}

\section{Experimental setup}

\subsection{Task}

We look at the task of image caption agreement, that is, given a visual scene and a statement, decide whether the latter is true for the former. See figure \ref{figure:examples} for some examples. The captions here are formal-semantics-style statements and not necessarily good descriptions, which is a vaguer concept and thus not as useful for evaluation. Instead, this task corresponds more to yes/no questions in VQA. Formulating the task as a binary choice problem is interesting from an evaluation perspective, as it allows for difficult \textit{minimally} incorrect instances \cite{Hodosh2016}.

\subsection{Datasets}

We generated various datasets based on existing ShapeWorld configurations. The different datasets are defined by the types of captions they contain. See figure \ref{figure:examples} for more details. Each dataset consists of 500k training instances, plus 10k validation and test instances. Training and validation scenes generally contain 1-4, 6-9 or 11-14 non-overlapping (unless mentioned otherwise) objects, further constrained depending on the dataset. Test scenes may in addition exhibit the withheld object numbers 5, 10 and 15, and contain withheld object types: \textit{``red square''}, \textit{``green triangle''}, \textit{``blue circle''}, \textit{``yellow rectangle''}, \textit{``magenta cross''}, \textit{``cyan ellipse''}. Consequently, the test data follows a slightly different distribution, where models are required to generalize to unseen object numbers and new attribute combinations to achieve a comparable score, similar to the CoGenT version of the CLEVR dataset\footnote{Note, however, that CLEVR CoGenT requires stronger generalization skills, as more shape-color combinations per shape/color are withheld.}.

\subsection{Models}

We focus on the FiLM model \cite{Perez2018} in this paper. The image is processed using a six-layer CNN (stride of 2 after the third and sixth layer) trained from scratch on the task. We found that the common approach of using a pretrained ResNet module did not perform well on our data. The caption as `question' is processed by a GRU. In four residual blocks, the processed image tensor is linearly modulated, conditioned on the caption embedding. Following global max-pooling, the classifier module produces the `answer', i.e.\ \textit{``true''} or \textit{``false''} in our case. We train the model for 100k iterations in all experiments, using the default hyperparameters. Training performance is measured on the validation set every 1k iterations for the first 10k iterations and every 5k afterwards. We also compare performance to two common baselines \cite{Johnson2017a} on selected datasets: CNN-LSTM and CNN-LSTM-SA. We will release the ShapeWorld-adapted FiLM repository and the generator configurations to create the datasets on acceptance of the paper.

\begin{figure*}
\centering
\resizebox{0.46\linewidth}{!}{
\begin{tikzpicture}
\begin{axis}[
every axis plot/.append style={line width=2.5pt},
cycle list name=Dark2,
mark options={mark=none},
width=10cm,
height=7.5cm,
xmin=0, xmax=100,
xtick={0,20,40,60,80,100},
xticklabel style={font=\footnotesize},
xmajorgrids,
ymin=0.49, ymax=1.0,
ytick={0.5,0.6,0.7,0.8,0.9,1.0},
ymajorgrids,
yticklabel style={font=\footnotesize},
grid style=dashed,
legend style={at={(1.0,0.5)},font=\scriptsize}
]
\addplot coordinates {
(0,0.498) (1,0.57) (2,0.562) (3,0.604) (4,0.617) (5,0.627) (6,0.642) (7,0.635) (8,0.668) (9,0.667) (10,0.685) (15,0.691) (20,0.726) (25,0.756) (30,0.812) (35,0.915) (40,0.988) (45,0.993) (50,0.995) (55,0.997) (60,0.997) (65,0.998) (70,0.999) (75,0.999) (80,0.999) (85,0.999) (90,0.999) (95,0.999) (100,1.0)
};
\addplot coordinates {
(0,0.499) (1,0.516) (2,0.521) (3,0.556) (4,0.576) (5,0.61) (6,0.618) (7,0.613) (8,0.611) (9,0.633) (10,0.64) (15,0.649) (20,0.652) (25,0.673) (30,0.691) (35,0.77) (40,0.836) (45,0.856) (50,0.862) (55,0.867) (60,0.87) (65,0.879) (70,0.93) (75,0.98) (80,0.991) (85,0.994) (90,0.99) (95,0.996) (100,0.997)
};
\addplot coordinates {
(0,0.49) (1,0.512) (2,0.525) (3,0.536) (4,0.525) (5,0.555) (6,0.547) (7,0.566) (8,0.58) (9,0.601) (10,0.599) (15,0.634) (20,0.657) (25,0.67) (30,0.695) (35,0.746) (40,0.808) (45,0.861) (50,0.917) (55,0.94) (60,0.943) (65,0.952) (70,0.966) (75,0.968) (80,0.97) (85,0.966) (90,0.971) (95,0.972) (100,0.985)
};
\addplot coordinates {
(0,0.498) (1,0.525) (2,0.528) (3,0.571) (4,0.549) (5,0.588) (6,0.547) (7,0.589) (8,0.602) (9,0.614) (10,0.61) (15,0.643) (20,0.658) (25,0.668) (30,0.683) (35,0.74) (40,0.794) (45,0.839) (50,0.885) (55,0.907) (60,0.931) (65,0.934) (70,0.943) (75,0.9) (80,0.956) (85,0.958) (90,0.964) (95,0.96) (100,0.969)
};
\addplot coordinates {
(0,0.49) (1,0.502) (2,0.507) (3,0.51) (4,0.494) (5,0.495) (6,0.527) (7,0.526) (8,0.544) (9,0.571) (10,0.562) (15,0.57) (20,0.606) (25,0.64) (30,0.666) (35,0.71) (40,0.76) (45,0.786) (50,0.791) (55,0.796) (60,0.81) (65,0.814) (70,0.839) (75,0.85) (80,0.863) (85,0.865) (90,0.868) (95,0.867) (100,0.87)
};
\addplot+[color=black!90,dashed] coordinates {
(0,0.49632000000000004) (1,0.52532) (2,0.52882) (3,0.55578) (4,0.5525) (5,0.5754400000000001) (6,0.57686) (7,0.58626) (8,0.60146) (9,0.6176999999999999) (10,0.61968) (15,0.63778) (20,0.66028) (25,0.6830400000000001) (30,0.7097000000000001) (35,0.77688) (40,0.8378) (45,0.86764) (50,0.8907599999999999) (55,0.9019) (60,0.9107800000000001) (65,0.91572) (70,0.9361200000000001) (75,0.9514400000000001) (80,0.9569799999999999) (85,0.9568) (90,0.95922) (95,0.95928) (100,0.9646999999999999)
};
\legend{existential,logical,numbers,quantifiers,relational,overall}
\end{axis}
\end{tikzpicture}}%
\hspace{0.04\linewidth}%
\resizebox{0.46\linewidth}{!}{
\begin{tikzpicture}
\begin{axis}[
every axis plot/.append style={line width=2.5pt},
cycle list name=Dark2,
mark options={mark=none},
width=10cm,
height=7.5cm,
xmin=0, xmax=100,
xtick={0,20,40,60,80,100},
xticklabel style={font=\footnotesize},
xmajorgrids,
ymin=0.49, ymax=1.0,
ytick={0.5,0.6,0.7,0.8,0.9,1.0},
ymajorgrids,
yticklabel style={font=\footnotesize},
grid style=dashed,
legend style={at={(1.0,0.95)},font=\scriptsize}
]
\addplot coordinates {
(0,0.501) (1,0.53) (2,0.545) (3,0.498) (4,0.53) (5,0.569) (6,0.54) (7,0.57) (8,0.57) (9,0.599) (10,0.582) (15,0.623) (20,0.628) (25,0.629) (30,0.635) (35,0.648) (40,0.629) (45,0.6) (50,0.648) (55,0.667) (60,0.626) (65,0.67) (70,0.685) (75,0.685) (80,0.694) (85,0.696) (90,0.7) (95,0.72) (100,0.732)
};
\addplot coordinates {
(0,0.5) (1,0.511) (2,0.516) (3,0.513) (4,0.515) (5,0.532) (6,0.537) (7,0.544) (8,0.536) (9,0.549) (10,0.557) (15,0.619) (20,0.625) (25,0.63) (30,0.623) (35,0.636) (40,0.631) (45,0.634) (50,0.649) (55,0.643) (60,0.587) (65,0.646) (70,0.652) (75,0.651) (80,0.657) (85,0.662) (90,0.663) (95,0.65) (100,0.666)
};
\addplot coordinates {
(0,0.509) (1,0.505) (2,0.516) (3,0.498) (4,0.525) (5,0.529) (6,0.512) (7,0.529) (8,0.529) (9,0.53) (10,0.551) (15,0.551) (20,0.55) (25,0.574) (30,0.582) (35,0.605) (40,0.582) (45,0.612) (50,0.607) (55,0.609) (60,0.61) (65,0.616) (70,0.629) (75,0.628) (80,0.63) (85,0.645) (90,0.646) (95,0.653) (100,0.66)
};
\addplot coordinates {
(0,0.501) (1,0.5) (2,0.523) (3,0.494) (4,0.534) (5,0.536) (6,0.536) (7,0.56) (8,0.555) (9,0.566) (10,0.585) (15,0.571) (20,0.584) (25,0.603) (30,0.599) (35,0.623) (40,0.608) (45,0.631) (50,0.632) (55,0.622) (60,0.629) (65,0.639) (70,0.641) (75,0.637) (80,0.651) (85,0.659) (90,0.65) (95,0.659) (100,0.67)
};
\addplot coordinates {
(0,0.506) (1,0.504) (2,0.504) (3,0.497) (4,0.506) (5,0.494) (6,0.493) (7,0.504) (8,0.506) (9,0.508) (10,0.517) (15,0.499) (20,0.511) (25,0.539) (30,0.528) (35,0.555) (40,0.526) (45,0.569) (50,0.575) (55,0.575) (60,0.579) (65,0.582) (70,0.598) (75,0.588) (80,0.6) (85,0.602) (90,0.604) (95,0.6) (100,0.612)
};
\pgfplotsset{cycle list shift=1}
\addplot coordinates {
(0,0.5) (1,0.498) (2,0.49) (3,0.493) (4,0.499) (5,0.49) (6,0.501) (7,0.504) (8,0.5) (9,0.5) (10,0.5) (15,0.502) (20,0.514) (25,0.526) (30,0.529) (35,0.541) (40,0.521) (45,0.54) (50,0.543) (55,0.54) (60,0.545) (65,0.55) (70,0.551) (75,0.545) (80,0.565) (85,0.572) (90,0.57) (95,0.609) (100,0.647)
};
\addplot+[color=black!90,dashed] coordinates {
(0,0.5032000000000001) (1,0.5095166666666666) (2,0.51765) (3,0.49948333333333333) (4,0.5186333333333333) (5,0.5269833333333334) (6,0.5202666666666667) (7,0.5370833333333334) (8,0.5331499999999999) (9,0.5426) (10,0.5494166666666667) (15,0.5613166666666667) (20,0.5701333333333333) (25,0.5839333333333333) (30,0.5832666666666667) (35,0.6017666666666667) (40,0.5835666666666667) (45,0.6075499999999999) (50,0.6095499999999999) (55,0.6100500000000001) (60,0.5969666666666666) (65,0.6176833333333333) (70,0.6264500000000001) (75,0.6229333333333333) (80,0.6333333333333334) (85,0.6399166666666667) (90,0.6414166666666666) (95,0.6513166666666667) (100,0.6666833333333334)
};
\legend{existential,logical,numbers,quantifiers,relational,superlatives,overall}
\end{axis}
\end{tikzpicture}}%
\par\vspace{0.4cm}%
\resizebox{0.46\linewidth}{!}{
\begin{tikzpicture}
\begin{axis}[
every axis plot/.append style={line width=2.5pt},
cycle list name=Dark2,
mark options={mark=none},
width=10cm,
height=7.5cm,
xmin=0, xmax=100,
xtick={0,20,40,60,80,100},
xticklabel style={font=\footnotesize},
xmajorgrids,
ymin=0.49, ymax=1.0,
ytick={0.5,0.6,0.7,0.8,0.9,1.0},
ymajorgrids,
yticklabel style={font=\footnotesize},
grid style=dashed,
legend style={at={(1.0,0.6)},font=\scriptsize}
]
\addplot coordinates {
(0,0.501) (1,0.554) (2,0.575) (3,0.524) (4,0.55) (5,0.54) (6,0.577) (7,0.586) (8,0.597) (9,0.61) (10,0.609) (15,0.617) (20,0.654) (25,0.689) (30,0.713) (35,0.74) (40,0.773) (45,0.846) (50,0.932) (55,0.974) (60,0.993) (65,0.996) (70,0.996) (75,0.997) (80,0.998) (85,0.997) (90,0.999) (95,0.998) (100,0.999)
};
\addplot coordinates {
(0,0.5) (1,0.52) (2,0.53) (3,0.507) (4,0.509) (5,0.517) (6,0.523) (7,0.506) (8,0.544) (9,0.575) (10,0.572) (15,0.63) (20,0.624) (25,0.646) (30,0.607) (35,0.655) (40,0.652) (45,0.711) (50,0.778) (55,0.819) (60,0.846) (65,0.847) (70,0.85) (75,0.854) (80,0.85) (85,0.856) (90,0.858) (95,0.858) (100,0.861)
};
\addplot coordinates {
(0,0.499) (1,0.521) (2,0.496) (3,0.53) (4,0.51) (5,0.507) (6,0.533) (7,0.515) (8,0.531) (9,0.558) (10,0.537) (15,0.565) (20,0.582) (25,0.622) (30,0.615) (35,0.639) (40,0.662) (45,0.664) (50,0.723) (55,0.782) (60,0.834) (65,0.888) (70,0.909) (75,0.943) (80,0.955) (85,0.944) (90,0.967) (95,0.958) (100,0.97)
};
\addplot coordinates {
(0,0.502) (1,0.537) (2,0.509) (3,0.539) (4,0.534) (5,0.527) (6,0.557) (7,0.517) (8,0.539) (9,0.57) (10,0.548) (15,0.594) (20,0.6) (25,0.629) (30,0.625) (35,0.645) (40,0.659) (45,0.683) (50,0.715) (55,0.77) (60,0.832) (65,0.86) (70,0.879) (75,0.896) (80,0.927) (85,0.934) (90,0.946) (95,0.948) (100,0.954)
};
\pgfplotsset{cycle list shift=1}
\addplot coordinates {
(0,0.5) (1,0.501) (2,0.497) (3,0.497) (4,0.504) (5,0.51) (6,0.49) (7,0.502) (8,0.5) (9,0.52) (10,0.502) (15,0.526) (20,0.538) (25,0.554) (30,0.569) (35,0.618) (40,0.647) (45,0.726) (50,0.82) (55,0.87) (60,0.903) (65,0.914) (70,0.923) (75,0.93) (80,0.938) (85,0.937) (90,0.94) (95,0.944) (100,0.945)
};
\addplot coordinates {
(0,0.5) (1,0.51) (2,0.503) (3,0.495) (4,0.501) (5,0.495) (6,0.499) (7,0.504) (8,0.505) (9,0.51) (10,0.5) (15,0.52) (20,0.535) (25,0.565) (30,0.571) (35,0.636) (40,0.693) (45,0.768) (50,0.858) (55,0.897) (60,0.924) (65,0.931) (70,0.934) (75,0.943) (80,0.95) (85,0.949) (90,0.949) (95,0.948) (100,0.952)
};
\addplot+[color=black!90,dashed] coordinates {
(0,0.5008833333333333) (1,0.5251333333333333) (2,0.5187666666666667) (3,0.51605) (4,0.5186833333333333) (5,0.5168499999999999) (6,0.5317) (7,0.5223833333333333) (8,0.53965) (9,0.5594833333333333) (10,0.5451) (15,0.5756833333333334) (20,0.59045) (25,0.61805) (30,0.6172333333333333) (35,0.6563) (40,0.6817833333333333) (45,0.7334999999999999) (50,0.8060833333333332) (55,0.8524833333333333) (60,0.8892500000000001) (65,0.9064000000000001) (70,0.9155166666666666) (75,0.92765) (80,0.9380833333333333) (85,0.9369999999999999) (90,0.9436833333333334) (95,0.9427) (100,0.9478333333333334)
};
\legend{existential,logical,numbers,quantifiers,implicit-rel,superlatives,overall}
\end{axis}
\end{tikzpicture}}
\hspace{0.04\linewidth}%
\resizebox{0.46\linewidth}{!}{
\begin{tikzpicture}
\begin{axis}[
every axis plot/.append style={line width=2.5pt},
cycle list name=Dark2,
mark options={mark=none},
width=10cm,
height=7.5cm,
xmin=0, xmax=100,
xtick={0,20,40,60,80,100},
xticklabel style={font=\footnotesize},
xmajorgrids,
ymin=0.49, ymax=1.0,
ytick={0.5,0.6,0.7,0.8,0.9,1.0},
ymajorgrids,
yticklabel style={font=\footnotesize},
grid style=dashed,
legend style={at={(1.0,0.8)},font=\scriptsize}
]
\addplot coordinates {
(0,0.498) (1,0.505) (2,0.542) (3,0.545) (4,0.501) (5,0.507) (6,0.528) (7,0.563) (8,0.549) (9,0.51) (10,0.499) (15,0.505) (20,0.532) (25,0.494) (30,0.504) (35,0.525) (40,0.522) (45,0.547) (50,0.552) (55,0.533) (60,0.592) (65,0.507) (70,0.543) (75,0.552) (80,0.531) (85,0.563) (90,0.596) (95,0.598) (100,0.57)
};
\addplot coordinates {
(0,0.499) (1,0.0) (2,0.514) (3,0.508) (4,0.501) (5,0.491) (6,0.513) (7,0.524) (8,0.512) (9,0.5) (10,0.506) (15,0.505) (20,0.508) (25,0.496) (30,0.508) (35,0.508) (40,0.498) (45,0.508) (50,0.508) (55,0.504) (60,0.539) (65,0.504) (70,0.514) (75,0.514) (80,0.521) (85,0.525) (90,0.555) (95,0.595) (100,0.521)
};
\addplot coordinates {
(0,0.49) (1,0.511) (2,0.509) (3,0.508) (4,0.509) (5,0.497) (6,0.51) (7,0.51) (8,0.51) (9,0.51) (10,0.493) (15,0.487) (20,0.498) (25,0.49) (30,0.507) (35,0.501) (40,0.503) (45,0.509) (50,0.511) (55,0.517) (60,0.525) (65,0.533) (70,0.515) (75,0.521) (80,0.514) (85,0.525) (90,0.54) (95,0.526) (100,0.552)
};
\addplot coordinates {
(0,0.498) (1,0.526) (2,0.501) (3,0.522) (4,0.501) (5,0.506) (6,0.514) (7,0.524) (8,0.526) (9,0.503) (10,0.504) (15,0.49) (20,0.533) (25,0.486) (30,0.506) (35,0.519) (40,0.503) (45,0.507) (50,0.533) (55,0.527) (60,0.54) (65,0.552) (70,0.531) (75,0.533) (80,0.535) (85,0.54) (90,0.571) (95,0.549) (100,0.566)
};
\addplot coordinates {
(0,0.49) (1,0.494) (2,0.5) (3,0.493) (4,0.506) (5,0.493) (6,0.491) (7,0.495) (8,0.497) (9,0.5) (10,0.492) (15,0.493) (20,0.501) (25,0.506) (30,0.5) (35,0.501) (40,0.5) (45,0.507) (50,0.5) (55,0.503) (60,0.505) (65,0.496) (70,0.508) (75,0.505) (80,0.503) (85,0.497) (90,0.512) (95,0.503) (100,0.501)
};
\addplot coordinates {
(0,0.49) (1,0.497) (2,0.5) (3,0.49) (4,0.504) (5,0.498) (6,0.502) (7,0.488) (8,0.498) (9,0.5) (10,0.496) (15,0.492) (20,0.49) (25,0.504) (30,0.5) (35,0.501) (40,0.5) (45,0.5) (50,0.5) (55,0.501) (60,0.0) (65,0.5) (70,0.51) (75,0.5) (80,0.5) (85,0.507) (90,0.513) (95,0.499) (100,0.501)
};
\addplot coordinates {
(0,0.499) (1,0.5) (2,0.5) (3,0.499) (4,0.5) (5,0.501) (6,0.498) (7,0.498) (8,0.502) (9,0.5) (10,0.496) (15,0.494) (20,0.499) (25,0.504) (30,0.498) (35,0.489) (40,0.5) (45,0.499) (50,0.496) (55,0.499) (60,0.502) (65,0.503) (70,0.502) (75,0.5) (80,0.499) (85,0.501) (90,0.502) (95,0.499) (100,0.5)
};
\addplot+[color=black!90,dashed] coordinates {
(0,0.49717142857142854) (1,0.5050428571428572) (2,0.5101714285714285) (3,0.5111571428571429) (4,0.5036857142857143) (5,0.4995142857142857) (6,0.5086) (7,0.5160571428571429) (8,0.5142) (9,0.5045428571428572) (10,0.4983285714285714) (15,0.4956714285714286) (20,0.5103142857142857) (25,0.49875714285714284) (30,0.5037571428571429) (35,0.507) (40,0.5042428571428571) (45,0.5115142857142858) (50,0.5157142857142857) (55,0.5124857142857143) (60,0.5294571428571428) (65,0.5141) (70,0.5181857142857142) (75,0.5183142857142857) (80,0.5151428571428571) (85,0.5242857142857142) (90,0.5421857142857143) (95,0.5388714285714286) (100,0.5316857142857142)
};
\legend{existential,logical,numbers,quantifiers,relational,implicit-rel,superlatives,overall}
\end{axis}
\end{tikzpicture}}
\vspace{-0.1cm}
\caption{
Performance per dataset of the FiLM model trained on broader combinations of datasets.}
\label{figure:performance2}
\end{figure*}
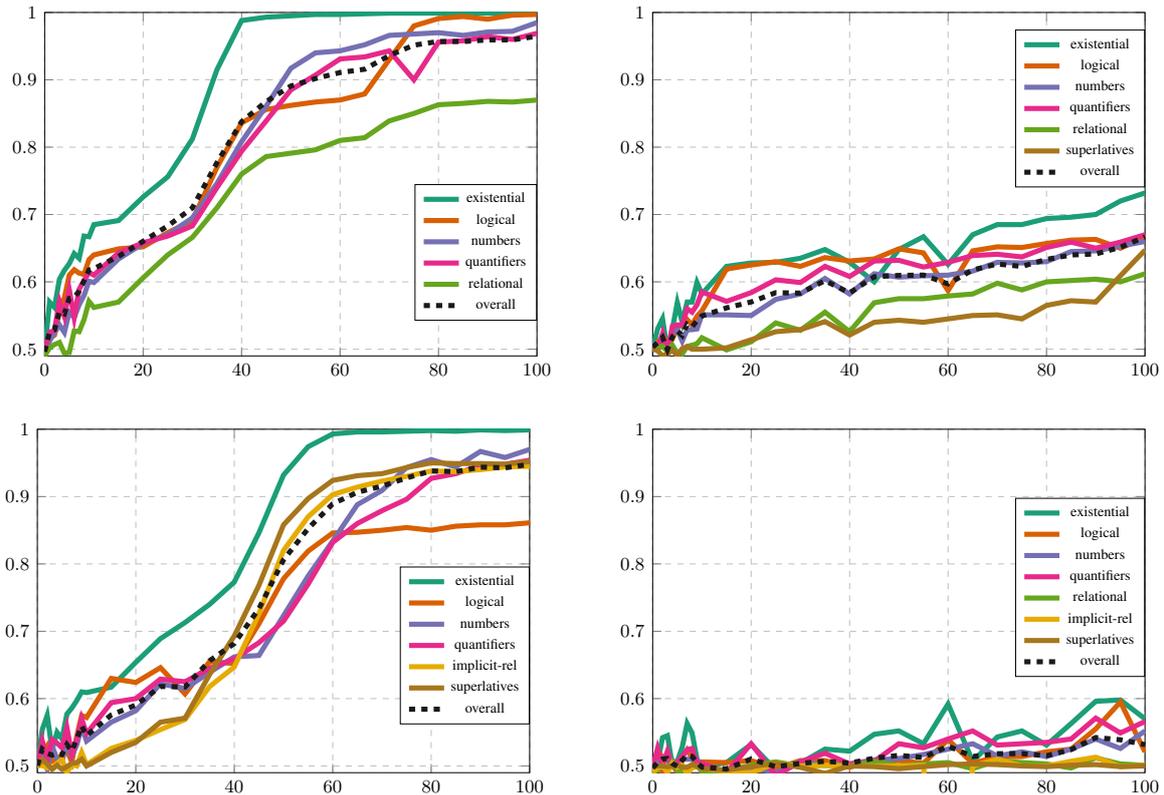

\section{Results}

\paragraph{Pretrained ResNet does not perform well.}
We started off experimenting with the FiLM default of using a pretrained ResNet module instead of a custom CNN. Versions with either a fixed or a trainable ResNet reach an accuracy of 65-70\% after 100k iterations on \dataset{existential}, which is substantially lower than our final result of 100\% (see appendix \ref{appendix:resnet}). This is surprisingly different from findings for CLEVR, where others previously reported the level of performance for either a pretrained ResNet or a custom CNN to be on a par \cite{Perez2018,Santoro2017}.

\paragraph{Overlapping objects can impede learning.}
Initially, we also did not explicitly configure data generation to prevent overlapping objects. This turned out to be a major obstacle for learning in some cases: while FiLM solves \dataset{existential} (99.7\%), performance on \dataset{numbers} stays at chance level (55.2\%). To investigate this further, we configured the generator to only permit a lower degree of overlap (default: 25\% max area overlap). 17.5\% shows no improvement, performance for 10\% is slightly better, but only in the case of 5\% area overlap we observe a substantially improved accuracy of $\sim$73\% after 100k iterations (see appendix \ref{appendix:overlap}). Since the background is black by default, we assume that having to learn to recognize objects contrasted with an unusual color can have a destructive influence on the overall learning process, at least in the case of more difficult tasks. Unless stated otherwise, we thus use overlap-free data for the following experiments, where number statements are learned perfectly.

\paragraph{Many datasets solved and simple generalization works.}
Overall, the FiLM model successfully learns many of our datasets (see figure \ref{figure:performance1}). \dataset{existential} is mastered after only 10k iterations and at the same speed as the trivial \dataset{single-shape} variant. \dataset{logical}, \dataset{numbers} and \dataset{quantifiers} reach close-to-perfect accuracy after around 60k iterations. The learning curves for these three tasks look remarkably alike and thus suggest a similar learning complexity for the model. Moreover, the FiLM model successfully generalizes to the test set in most cases (see figure \ref{figure:performance1}), showing that it learned the ability for simple compositional recombination. Only for the simplified variants \dataset{single-shape} and \dataset{simple-spatial} test performance is markedly lower, indicating that there is not enough incentive to learn a compositional representation here. This is presumably because their simplicity makes overfitting a feasible option, due to the lack of distractors which may require to distinguish individual attributes.

\paragraph{Stacked attention is not consistently superior.}
We investigated the performance of two common baselines, CNN-LSTM and CNN-LSTM-SA (see figure \ref{figure:performance1} as well as appendix \ref{appendix:cnnlstm} and \ref{appendix:cnnlstmsa}). While FiLM consistently outperforms both baselines as expected, the supposedly superior CNN-LSTM-SA \cite{Yang2016,Johnson2017a} does not always improve upon the results of CNN-LSTM. However, CNN-LSTM-SA in some cases shows stronger generalization to the test distribution, whereas performance always drops for CNN-LSTM. We note, though, that it is unclear whether the ability to generalize is expected of a system that does not fully solve a task to begin with.

\paragraph{Failure to learn relational statements.}
Surprisingly, we find that, with the exception of \dataset{simple-spatial}, FiLM struggles to learn anything when trained on the various datasets requiring some form of relational reasoning (see figure \ref{figure:performance1}): \dataset{relational}, \dataset{implicit-relational} and \dataset{superlatives} (referred to as \dataset{relational-like} below). We also tried subsets of relations in \dataset{relational} (e.g., only spatial relations), with the same result. The only exception is the simplistic two-object variant \dataset{simple-spatial}. But even here, learning is comparatively slow and, after plateauing for around 50k iterations at $\sim$75\%, reaches only $\sim$85\% after 100k iterations (similar to baselines, although the curve indicates that performance is still improving). This further emphasizes the complexity for FiLM to learn relational statements.

\paragraph{Training on a broader set of instances.}
Datasets like CLEVR consist of a mix of instance types requiring different understanding abilities. Our assumption is that the simpler instances help to stabilize and guide the overall learning process, so that the more complex instances are also learned eventually\footnote{When referring to ``simple'' and ``complex'' or ``difficult'' instances here and in the following, we always mean with respect to the ability of the FiLM model to learn these instances.}, hence models are able to achieve close-to-perfect performance overall. We tested this assumption by training on a broader mix of \dataset{existential}, \dataset{logical}, \dataset{numbers}, \dataset{quantifiers} in combination with some of the \dataset{relational-like} datasets (see figure \ref{figure:performance2} and appendix \ref{appendix:mixers}). Indeed, FiLM is able to successfully learn mixer datasets involving one of the more difficult datasets, or two in the case of \dataset{implicit-relational} and \dataset{superlatives}.

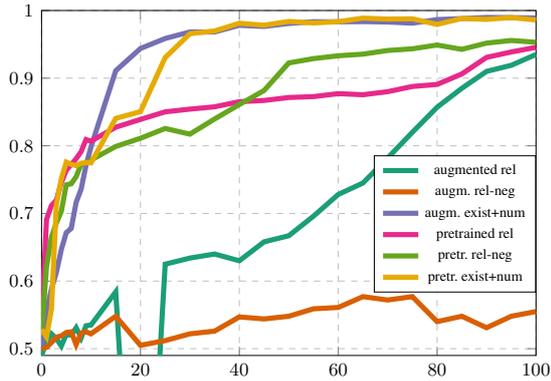
\begin{figure}
\resizebox{0.96\linewidth}{!}{
\begin{tikzpicture}
\begin{axis}[
every axis plot/.append style={line width=2.5pt},
cycle list name=Dark2,
mark options={mark=none},
width=10cm,
height=7.5cm,
xmin=0, xmax=100,
xtick={0,20,40,60,80,100},
xticklabel style={font=\footnotesize},
xmajorgrids,
ymin=0.49, ymax=1.0,
ytick={0.5,0.6,0.7,0.8,0.9,1.0},
ymajorgrids,
yticklabel style={font=\footnotesize},
grid style=dashed,
legend style={at={(1.0,0.58)},font=\scriptsize}
]
\addplot coordinates {
(0,0.49) (1,0.51) (2,0.522) (3,0.515) (4,0.504) (5,0.521) (6,0.52) (7,0.531) (8,0.516) (9,0.534) (10,0.535) (15,0.584) (20,0.0) (25,0.625) (30,0.634) (35,0.64) (40,0.63) (45,0.658) (50,0.667) (55,0.696) (60,0.728) (65,0.745) (70,0.782) (75,0.82) (80,0.857) (85,0.885) (90,0.91) (95,0.919) (100,0.935)
};
\addplot coordinates {
(0,0.5) (1,0.502) (2,0.509) (3,0.517) (4,0.519) (5,0.524) (6,0.525) (7,0.506) (8,0.523) (9,0.525) (10,0.522) (15,0.548) (20,0.505) (25,0.512) (30,0.522) (35,0.526) (40,0.547) (45,0.544) (50,0.548) (55,0.559) (60,0.561) (65,0.577) (70,0.572) (75,0.577) (80,0.54) (85,0.548) (90,0.531) (95,0.548) (100,0.555)
};
\addplot coordinates {
(0,0.5038) (1,0.5565) (2,0.58805) (3,0.6128) (4,0.6467) (5,0.6716500000000001) (6,0.6778) (7,0.7164999999999999) (8,0.73585) (9,0.77125) (10,0.7985) (15,0.911) (20,0.9438500000000001) (25,0.95865) (30,0.9682999999999999) (35,0.96825) (40,0.9781500000000001) (45,0.97645) (50,0.98085) (55,0.9834499999999999) (60,0.9833000000000001) (65,0.9836) (70,0.983) (75,0.9813000000000001) (80,0.9866) (85,0.98805) (90,0.98975) (95,0.98955) (100,0.98895)
};
\addplot coordinates {
(0,0.5354) (1,0.691) (2,0.7117) (3,0.7196) (4,0.7455) (5,0.7638) (6,0.7724) (7,0.7809) (8,0.7913) (9,0.8092) (10,0.8068) (15,0.8275) (20,0.839) (25,0.8502) (30,0.8542) (35,0.8576) (40,0.8652) (45,0.867) (50,0.8714) (55,0.8727) (60,0.8771) (65,0.8755) (70,0.88) (75,0.8877) (80,0.8907) (85,0.9062) (90,0.9307) (95,0.9387) (100,0.9456)
};
\addplot coordinates {
(0,0.5233) (1,0.621) (2,0.6653) (3,0.686) (4,0.703) (5,0.7423) (6,0.7438) (7,0.7543) (8,0.7728) (9,0.7748) (10,0.7789) (15,0.799) (20,0.8111) (25,0.8256) (30,0.8172) (35,0.8396) (40,0.861) (45,0.8817) (50,0.9225) (55,0.9289) (60,0.9331) (65,0.9355) (70,0.9409) (75,0.9434) (80,0.949) (85,0.9426) (90,0.9518) (95,0.9557) (100,0.953)
};
\addplot coordinates {
(0,0.5285) (1,0.5175000000000001) (2,0.5583) (3,0.7132000000000001) (4,0.75125) (5,0.7758499999999999) (6,0.7725500000000001) (7,0.7706500000000001) (8,0.77425) (9,0.775) (10,0.7746999999999999) (15,0.8405) (20,0.8504499999999999) (25,0.93045) (30,0.9657) (35,0.9697) (40,0.98115) (45,0.9782500000000001) (50,0.98385) (55,0.9816) (60,0.9835) (65,0.98875) (70,0.98735) (75,0.98775) (80,0.9794) (85,0.9883) (90,0.9869) (95,0.9895499999999999) (100,0.986)
};
\legend{augmented rel,augm.\ rel-neg,augm.\ exist+num,pretrained rel,pretr.\ rel-neg,pretr.\ exist+num}
\end{axis}
\end{tikzpicture}}
\caption{
Performance on \dataset{relational/-negation} or \dataset{existential+numbers} (with overlap), when augmented with / pretrained on \dataset{simple-spatial} or \dataset{existential} instances, respectively.}
\label{figure:performance3}
\end{figure}

\paragraph{Augmenting with a simpler dataset.}
Additionally, we looked at the situation of a complex dataset paired with a simpler one, where instances of the latter can act as `pedagogical' examples of a more general instance type (see figure \ref{figure:performance3}). First, the FiLM model reaches $\sim$95\% accuracy on a dataset augmenting the complex \dataset{relational} with the simple \dataset{simple-spatial} dataset. Second, in the failure case of \dataset{numbers} with overlapping objects, training on a combination with \dataset{existential} instances (with overlap) helps the model to also solve instances of the former.

\paragraph{Improvements by mixing/augmenting are unstable.}
Further investigation reveals that this `synergy effect' of combining different datasets is very sensitive to the composition of the training set. On the one hand, FiLM fails to learn most mixer datasets with two or more \dataset{relational-like} components (see figure \ref{figure:performance2} and appendix \ref{appendix:mixers}). These results further indicate that \dataset{relational} seems to be the most complex of the \dataset{relational-like} datasets. On the other hand, even a slightly unbalanced distribution of 45\% or 60\% \dataset{simple-spatial} instances with 55\% or 40\% \dataset{relational} shows no improvement above chance level (see appendix \ref{appendix:distribution}). Moreover, instead of skewing the distribution, performance also stagnates when training on a combination with the more complex \dataset{relational-negation} instead of its negation-free variant (see figure \ref{figure:performance3}).

\paragraph{The effectiveness of pretraining.}
In another series of experiments we investigated whether pretraining on simpler instances can bootstrap a successful learning process on more complex datasets, which is the assumption underlying curriculum learning \cite{Elman1993,Bengio2009}. For this, we take the model trained for 100k iterations on \dataset{simple-spatial} and apply it to other \dataset{relational-like} datasets (see figure \ref{figure:performance3}). For both \dataset{relational} as well as \dataset{relational-negation} we observe a sharp increase in performance at the start, reaching $\sim$95\% accuracy after 100k iterations. We particularly want to draw attention to the fact that the pretrained model reaches and eventually surpasses its previous performance level of $\sim$85\% after only 20k/40k iterations, despite the more complex instances. Note also that the model trained on \dataset{relational-negation} at some point seems to benefit from this dataset's increased complexity. Finally, we also confirmed that, in the case of overlapping objects, the system pretrained on \dataset{existential} can subsequently also be trained to learn added \dataset{numbers} instances (see figure \ref{figure:performance3}).

\paragraph{Differences to findings for CLEVR}
\begin{itemize}[leftmargin=0.45cm,topsep=0.05cm,itemsep=0.05cm,parsep=0cm]
\item Pretrained ResNet does not perform well.
\item Overlapping objects can impede learning.
\item Simple compositional generalization (simpler than CLEVR CoGenT) is learned perfectly.
\item Relational statements are substantially more difficult to learn, at least in isolation.
\item The presence of simpler instances likely benefits the learning of more complex ones.
\item Performance on CLEVR does not transfer to all kinds of `CLEVR-like' abstract data.
\end{itemize}

\section{Related work}

Besides ShapeWorld and CLEVR, there is a number of other abstract VQA-like datasets, most notably, the NLVR \cite{Suhr2017} and the COG dataset \cite{Yang2018}. Of these, COG is most similar to ShapeWorld in its explicit focus on providing a test platform for a variety of tasks, while NLVR uses crowdsourced captions and consequently makes controlling for certain instance types more difficult. Other examples include SHAPES \cite{Andreas2016a} and Sort-of-CLEVR \cite{Santoro2017}, both of which act as proofs of concept in the respective paper.

Automatically generated language(-like) data is sometimes used to analyze the algorithmic capabilities of neural network models to efficiently process data of a certain structure. From early investigations into the ability of LSTMs to handle various formal grammars \cite{Gers2001}, to an analysis of stack-augmented RNNs \cite{Joulin2015}, to recently published negative findings on the compositional skills of sequence-to-sequence models \cite{Lake2017}. Like ShapeWorld, the bAbI test suite \cite{Weston2015} is an example of a more general and task-focused evaluation platform, using synthetic data for a range of targeted subtasks.

An alternative to automatically generating abstract data is to automatically modifying real-world datasets in a systematic way, with regard to evaluating a model's ability to spot invalid alterations. This can be seen as a form of `lightly' artificial data for evaluation purposes. \newcite{Hodosh2016} investigate image captioning models by swapping, replacing or removing noun phrases. Similarly, \newcite{Shekhar2017} replace nouns based on semantically related but incorrect words. \newcite{Jia2017}, in contrast, insert adversarially chosen distractor sentences into reading comprehension problems.

Besides multiple examples of a state-of-the-art model with surprisingly low performance on such diagnostic datasets/modifications, other recent findings emphasize the need for a more thorough analysis of existing systems and results. On the one hand, there is a range of papers showing competitive performance of simple, sometimes trivial, baseline systems for supposedly difficult benchmark datasets \cite{Poliak2018,Merity2018}. On the other hand, attempts to replicate experiments and large-scale comparisons of extensively tuned systems reveal the brittleness of many reported results/improvements \cite{Melis2018,Lucic2017,Henderson2018}.

\section{Discussion and conclusion}

We have shown how the FiLM model is not able to learn to correctly understand relational statements when trained on a dataset of such statements only. Furthermore, we have investigated two mechanisms which help alleviate these difficulties: augmenting training data with instances that are easier to learn, and pretraining on such simpler instances before moving to more complex ones. The first approach turns out to be very sensitive to the precise composition of the training set, while the second one leads to more robust improvements in our experiments.

In essence, mixing instances ultimately results in big all-encompassing datasets for general tasks like VQA, where a variety of skills is assumed to be learned implicitly from a lot of input-output pairs. While our results confirm that this is possible (at least for synthetic data), they strongly question the robustness of this process. We showed how otherwise successful learning breaks down when the combined dataset is too complex or the mixing distribution is chosen wrongly. We emphasize that these findings are based on clean and controlled abstract data, while the situation is even more complex for real-world datasets.

Such sensitivity of the learning process to structural details of the training data is usually not considered, but might be able to explain some of the instability effects that are generally attributed to hyperparameter choice, random seeds, etc. Since it is hard to conceive how real-world data could ever be controlled to the degree possible with synthetic data, researchers should be more skeptical of complex architectures for only a single dataset, and instead encourage the reporting of negative instability/transferability results.

Our findings resulted from a careful in-depth analysis of a single model on a range of instance types and configurations, as opposed to a single dataset -- even an explicitly diagnostic one, like CLEVR. This motivates our recommendation to abandon the idea of \textit{`datasets as tasks'}, and to shift focus from model building to model analysis. As a way forward, our findings suggest the potential of curriculum learning as a more robust and effective alternative to bigger monolithic datasets.

\appendix

\section{Learning curves for other experiments}

\subsection{Pretrained ResNet module}\label{appendix:resnet}

\begin{center}
\resizebox{0.96\linewidth}{!}{
\begin{tikzpicture}
\begin{axis}[
every axis plot/.append style={line width=2.5pt},
cycle list name=Dark2,
mark options={mark=none},
width=10cm,
height=7.5cm,
xmin=0, xmax=100,
xtick={0,20,40,60,80,100},
xticklabel style={font=\footnotesize},
xmajorgrids,
ymin=0.49, ymax=1.0,
ytick={0.5,0.6,0.7,0.8,0.9,1.0},
ymajorgrids,
yticklabel style={font=\footnotesize},
grid style=dashed,
legend style={at={(1.0,0.9)},font=\scriptsize}
]
\addplot coordinates {
(0,0.4991) (1,0.5837) (2,0.5988) (3,0.599) (4,0.6058) (5,0.61595) (6,0.6206) (7,0.6202) (8,0.61675) (9,0.60605) (10,0.61745) (15,0.61125) (20,0.63595) (25,0.64185) (30,0.64885) (35,0.64365) (40,0.64845) (45,0.65465) (50,0.66545) (55,0.66405) (60,0.6613) (65,0.67075) (70,0.6643) (75,0.67435) (80,0.673) (85,0.67485) (90,0.6781) (95,0.66695) (100,0.67705)
};
\addplot coordinates {
(0,0.5009) (1,0.5912) (2,0.6123) (3,0.6049) (4,0.61335) (5,0.6178) (6,0.6139) (7,0.6219) (8,0.5896) (9,0.63035) (10,0.62335) (15,0.6519) (20,0.6774) (25,0.6811) (30,0.68155) (35,0.68365) (40,0.6815) (45,0.68555) (50,0.6851) (55,0.6847) (60,0.6862) (65,0.6868) (70,0.68485) (75,0.6825) (80,0.68755) (85,0.68765) (90,0.68545) (95,0.6848) (100,0.68625)
};
\addplot coordinates {
(0,0.49255) (1,0.5521) (2,0.54945) (3,0.554) (4,0.56055) (5,0.557) (6,0.55535) (7,0.56445) (8,0.548) (9,0.5546) (10,0.569) (15,0.5543) (20,0.56095) (25,0.56495) (30,0.5787) (35,0.57675) (40,0.57345) (45,0.5799) (50,0.57375) (55,0.57275) (60,0.5741) (65,0.5757) (70,0.5723) (75,0.5796) (80,0.584) (85,0.5077) (90,0.6016) (95,0.5587) (100,0.6076)
};
\addplot coordinates {
(0,0.49255) (1,0.55165) (2,0.55265) (3,0.5595) (4,0.54965) (5,0.55855) (6,0.56985) (7,0.56245) (8,0.5657) (9,0.57685) (10,0.5708) (15,0.6027) (20,0.6058) (25,0.60545) (30,0.6085) (35,0.6124) (40,0.61565) (45,0.6131) (50,0.6204) (55,0.62075) (60,0.65415) (65,0.6568) (70,0.65855) (75,0.66445) (80,0.6651) (85,0.66765) (90,0.66945) (95,0.6704) (100,0.66345)
};
\addplot coordinates {
(0,0.4942) (1,0.51115) (2,0.5105) (3,0.51755) (4,0.4992) (5,0.49495) (6,0.5111) (7,0.50655) (8,0.4973) (9,0.50625) (10,0.49695) (15,0.50675) (20,0.5049) (25,0.50575) (30,0.50245) (35,0.4987) (40,0.50185) (45,0.4978) (50,0.4942) (55,0.49435) (60,0.50355) (65,0.4975) (70,0.49955) (75,0.50705) (80,0.5058) (85,0.51535) (90,0.51575) (95,0.5093) (100,0.4978)
};
\addplot coordinates {
(0,0.4942) (1,0.5134) (2,0.499) (3,0.5098) (4,0.5128) (5,0.5097) (6,0.5064) (7,0.4941) (8,0.50115) (9,0.5077) (10,0.50125) (15,0.5057) (20,0.5048) (25,0.5089) (30,0.5182) (35,0.49735) (40,0.5055) (45,0.49425) (50,0.5058) (55,0.51205) (60,0.5187) (65,0.51135) (70,0.49745) (75,0.4977) (80,0.4942) (85,0.49745) (90,0.5065) (95,0.51035) (100,0.5029)
};
\legend{existential fixed,existential trainable,numbers fixed,numbers trainable,relational fixed,relational trainable}
\end{axis}
\end{tikzpicture}}
\end{center}

\subsection{Overlapping objects}\label{appendix:overlap}

\begin{center}
\resizebox{0.96\linewidth}{!}{
\begin{tikzpicture}
\begin{axis}[
every axis plot/.append style={line width=2.5pt},
cycle list name=Dark2,
mark options={mark=none},
width=10cm,
height=7.5cm,
xmin=0, xmax=100,
xtick={0,20,40,60,80,100},
xticklabel style={font=\footnotesize},
xmajorgrids,
ymin=0.49, ymax=1.0,
ytick={0.5,0.6,0.7,0.8,0.9,1.0},
ymajorgrids,
yticklabel style={font=\footnotesize},
grid style=dashed,
legend style={at={(1.0,0.9)},font=\scriptsize}
]
\addplot coordinates {
(0,0.5093) (1,0.5342) (2,0.5491) (3,0.5624) (4,0.548) (5,0.5685) (6,0.5657) (7,0.5785) (8,0.5872) (9,0.5686) (10,0.5855) (15,0.6223) (20,0.6983) (25,0.7234) (30,0.7235) (35,0.759) (40,0.861) (45,0.9019) (50,0.9578) (55,0.9664) (60,0.9838) (65,0.9897) (70,0.9909) (75,0.9915) (80,0.9946) (85,0.9879) (90,0.9915) (95,0.9923) (100,0.9959)
};
\addplot coordinates {
(0,0.4942) (1,0.5561) (2,0.5296) (3,0.5446) (4,0.5549) (5,0.555) (6,0.5406) (7,0.5586) (8,0.5616) (9,0.5604) (10,0.5579) (15,0.5684) (20,0.5668) (25,0.5652) (30,0.5713) (35,0.569) (40,0.5564) (45,0.5743) (50,0.6054) (55,0.6308) (60,0.6489) (65,0.6471) (70,0.6585) (75,0.6593) (80,0.6633) (85,0.6692) (90,0.6661) (95,0.6779) (100,0.726)
};
\addplot coordinates {
(0,0.4957) (1,0.5406) (2,0.5497) (3,0.5386) (4,0.5454) (5,0.5548) (6,0.5472) (7,0.5435) (8,0.5529) (9,0.5516) (10,0.5527) (15,0.5528) (20,0.5188) (25,0.5462) (30,0.535) (35,0.5535) (40,0.5429) (45,0.557) (50,0.5819) (55,0.593) (60,0.5964) (65,0.5993) (70,0.6004) (75,0.6075) (80,0.604) (85,0.6069) (90,0.6041) (95,0.6038) (100,0.6082)
};
\addplot coordinates {
(0,0.4962) (1,0.5131) (2,0.5477) (3,0.5431) (4,0.5481) (5,0.5402) (6,0.5553) (7,0.5548) (8,0.5332) (9,0.5528) (10,0.5525) (15,0.539) (20,0.5422) (25,0.5533) (30,0.5558) (35,0.5574) (40,0.5388) (45,0.5414) (50,0.5608) (55,0.5574) (60,0.5486) (65,0.5416) (70,0.5558) (75,0.5476) (80,0.5351) (85,0.5602) (90,0.5532) (95,0.5485) (100,0.5533)
};
\addplot coordinates {
(0,0.5046) (1,0.547) (2,0.5583) (3,0.5511) (4,0.5283) (5,0.55) (6,0.5452) (7,0.5509) (8,0.534) (9,0.5335) (10,0.5512) (15,0.5542) (20,0.5597) (25,0.5248) (30,0.5542) (35,0.5425) (40,0.5563) (45,0.5408) (50,0.556) (55,0.5377) (60,0.5545) (65,0.5612) (70,0.5587) (75,0.5615) (80,0.5576) (85,0.551) (90,0.5529) (95,0.559) (100,0.5519)
};
\addplot coordinates {
(0,0.4977) (1,0.6777) (2,0.7451) (3,0.7621) (4,0.8361) (5,0.8793) (6,0.9314) (7,0.958) (8,0.9582) (9,0.9797) (10,0.9832) (15,0.9868) (20,0.9934) (25,0.9951) (30,0.9963) (35,0.9951) (40,0.9973) (45,0.9962) (50,0.9968) (55,0.9953) (60,0.9957) (65,0.9975) (70,0.9975) (75,0.9976) (80,0.9942) (85,0.9978) (90,0.997) (95,0.9969) (100,0.9973)
};
\legend{overlap-free,5\% numbers,10\% numbers,17.5\% numbers,25\% numbers,25\% existential}
\end{axis}
\end{tikzpicture}}
\end{center}

\subsection{CNN-LSTM baseline}\label{appendix:cnnlstm}

\begin{center}
\resizebox{0.96\linewidth}{!}{
\begin{tikzpicture}
\begin{axis}[
every axis plot/.append style={line width=2.5pt},
cycle list name=Dark2,
mark options={mark=none},
width=10cm,
height=7.5cm,
xmin=0, xmax=100,
xtick={0,20,40,60,80,100},
xticklabel style={font=\footnotesize},
xmajorgrids,
ymin=0.49, ymax=1.0,
ytick={0.5,0.6,0.7,0.8,0.9,1.0},
ymajorgrids,
yticklabel style={font=\footnotesize},
grid style=dashed,
legend style={at={(1.0,0.37)},font=\scriptsize}
]
\addplot coordinates {
(0,0.5029) (1,0.6425) (2,0.6857) (3,0.6947) (4,0.6995) (5,0.7059) (6,0.7116) (7,0.7121) (8,0.7136) (9,0.715) (10,0.7267) (15,0.8243) (20,0.8831) (25,0.9102) (30,0.942) (35,0.9608) (40,0.9735) (45,0.9852) (50,0.9881) (55,0.9913) (60,0.9903) (65,0.9982) (70,0.9981) (75,0.9938) (80,0.9988) (85,0.9989) (90,0.999) (95,0.9988) (100,0.9995)
};
\addplot coordinates {
(0,0.4992) (1,0.5513) (2,0.5675) (3,0.6214) (4,0.6275) (5,0.634) (6,0.6354) (7,0.6485) (8,0.6496) (9,0.6549) (10,0.655) (15,0.6613) (20,0.6732) (25,0.685) (30,0.5961) (35,0.6584) (40,0.6907) (45,0.6948) (50,0.6985) (55,0.7059) (60,0.72) (65,0.7278) (70,0.746) (75,0.7582) (80,0.7695) (85,0.7779) (90,0.7804) (95,0.7904) (100,0.7973)
};
\addplot coordinates {
(0,0.5004) (1,0.552) (2,0.5605) (3,0.555) (4,0.5643) (5,0.5686) (6,0.5611) (7,0.5403) (8,0.559) (9,0.5305) (10,0.5383) (15,0.5637) (20,0.5868) (25,0.6009) (30,0.6407) (35,0.679) (40,0.6821) (45,0.6989) (50,0.7045) (55,0.7121) (60,0.714) (65,0.7229) (70,0.7173) (75,0.7296) (80,0.7316) (85,0.7358) (90,0.747) (95,0.747) (100,0.7499)
};
\addplot coordinates {
(0,0.5001) (1,0.5768) (2,0.5779) (3,0.588) (4,0.5961) (5,0.6219) (6,0.6223) (7,0.6218) (8,0.6321) (9,0.6238) (10,0.6239) (15,0.5321) (20,0.5679) (25,0.5937) (30,0.5866) (35,0.6291) (40,0.6505) (45,0.6822) (50,0.6278) (55,0.6614) (60,0.6796) (65,0.6943) (70,0.6973) (75,0.7164) (80,0.6897) (85,0.7281) (90,0.7221) (95,0.721) (100,0.7213)
};
\pgfplotsset{cycle list shift=3}
\addplot coordinates {
(0,0.5005) (1,0.4926) (2,0.5723) (3,0.6427) (4,0.642) (5,0.6395) (6,0.643) (7,0.6475) (8,0.6444) (9,0.6559) (10,0.6614) (15,0.6426) (20,0.5494) (25,0.6246) (30,0.5056) (35,0.6916) (40,0.5058) (45,0.5056) (50,0.7167) (55,0.5852) (60,0.5799) (65,0.7146) (70,0.6958) (75,0.7375) (80,0.7532) (85,0.7674) (90,0.8012) (95,0.787) (100,0.814)
};
\legend{existential,logical,numbers,quantifiers,simple-spatial}
\end{axis}
\end{tikzpicture}}
\end{center}

\subsection{CNN-LSTM-SA baseline}\label{appendix:cnnlstmsa}

\begin{center}
\resizebox{0.96\linewidth}{!}{
\begin{tikzpicture}
\begin{axis}[
every axis plot/.append style={line width=2.5pt},
cycle list name=Dark2,
mark options={mark=none},
width=10cm,
height=7.5cm,
xmin=0, xmax=100,
xtick={0,20,40,60,80,100},
xticklabel style={font=\footnotesize},
xmajorgrids,
ymin=0.49, ymax=1.0,
ytick={0.5,0.6,0.7,0.8,0.9,1.0},
ymajorgrids,
yticklabel style={font=\footnotesize},
grid style=dashed,
legend style={at={(1.0,0.37)},font=\scriptsize}
]
\addplot coordinates {
(0,0.5042) (1,0.6833) (2,0.6789) (3,0.7147) (4,0.7146) (5,0.7283) (6,0.7731) (7,0.8136) (8,0.8757) (9,0.9047) (10,0.9446) (15,0.9873) (20,0.9916) (25,0.9814) (30,0.9972) (35,0.998) (40,0.9962) (45,0.9993) (50,0.9995) (55,0.9989) (60,0.9994) (65,0.9997) (70,0.9992) (75,0.9994) (80,0.9996) (85,0.9995) (90,0.9996) (95,0.9995) (100,0.9996)
};
\addplot coordinates {
(0,0.497) (1,0.5975) (2,0.6303) (3,0.6194) (4,0.637) (5,0.6383) (6,0.6389) (7,0.6386) (8,0.6415) (9,0.6357) (10,0.6482) (15,0.652) (20,0.6596) (25,0.6597) (30,0.6703) (35,0.6734) (40,0.6822) (45,0.6972) (50,0.6941) (55,0.7012) (60,0.7007) (65,0.7034) (70,0.7061) (75,0.712) (80,0.6284) (85,0.7337) (90,0.7414) (95,0.7478) (100,0.765)
};
\addplot coordinates {
(0,0.4906) (1,0.5529) (2,0.5807) (3,0.5712) (4,0.5731) (5,0.5729) (6,0.5789) (7,0.5853) (8,0.6029) (9,0.6021) (10,0.603) (15,0.615) (20,0.6172) (25,0.657) (30,0.6542) (35,0.6852) (40,0.7281) (45,0.7419) (50,0.7821) (55,0.8452) (60,0.8821) (65,0.9028) (70,0.9036) (75,0.914) (80,0.9724) (85,0.9782) (90,0.977) (95,0.9848) (100,0.9908)
};
\addplot coordinates {
(0,0.4982) (1,0.563) (2,0.5765) (3,0.5719) (4,0.5723) (5,0.5841) (6,0.5883) (7,0.5807) (8,0.561) (9,0.595) (10,0.6148) (15,0.6197) (20,0.6213) (25,0.6293) (30,0.6312) (35,0.6613) (40,0.6736) (45,0.689) (50,0.6996) (55,0.7204) (60,0.7388) (65,0.7585) (70,0.7956) (75,0.8074) (80,0.812) (85,0.8247) (90,0.8271) (95,0.8421) (100,0.8478)
};
\pgfplotsset{cycle list shift=3}
\addplot coordinates {
(0,0.4981) (1,0.5459) (2,0.5652) (3,0.5547) (4,0.583) (5,0.5957) (6,0.57) (7,0.6026) (8,0.6041) (9,0.5996) (10,0.6197) (15,0.6641) (20,0.6795) (25,0.5585) (30,0.6843) (35,0.5532) (40,0.6931) (45,0.6971) (50,0.6963) (55,0.6968) (60,0.7035) (65,0.7255) (70,0.735) (75,0.7069) (80,0.739) (85,0.7724) (90,0.7836) (95,0.7984) (100,0.8191)
};
\legend{existential,logical,numbers,quantifiers,simple-spatial}
\end{axis}
\end{tikzpicture}}
\end{center}

\subsection{Broader combinations of datasets}\label{appendix:mixers}

\begin{center}
\resizebox{0.96\linewidth}{!}{
\begin{tikzpicture}
\begin{axis}[
every axis plot/.append style={line width=2.5pt},
cycle list name=Dark2,
mark options={mark=none},
width=10cm,
height=7.5cm,
xmin=0, xmax=100,
xtick={0,20,40,60,80,100},
xticklabel style={font=\footnotesize},
xmajorgrids,
ymin=0.49, ymax=1.0,
ytick={0.5,0.6,0.7,0.8,0.9,1.0},
ymajorgrids,
yticklabel style={font=\footnotesize},
grid style=dashed,
legend style={at={(1.0,0.5)},font=\scriptsize}
]
\addplot coordinates {
(0,0.498) (1,0.568) (2,0.584) (3,0.589) (4,0.611) (5,0.612) (6,0.612) (7,0.604) (8,0.625) (9,0.63) (10,0.649) (15,0.711) (20,0.767) (25,0.903) (30,0.97) (35,0.996) (40,0.99) (45,0.998) (50,0.998) (55,0.998) (60,0.999) (65,0.999) (70,0.999) (75,0.993) (80,0.999) (85,0.999) (90,0.998) (95,0.997) (100,0.999)
};
\addplot coordinates {
(0,0.499) (1,0.528) (2,0.551) (3,0.528) (4,0.633) (5,0.62) (6,0.59) (7,0.612) (8,0.627) (9,0.639) (10,0.627) (15,0.65) (20,0.661) (25,0.742) (30,0.8) (35,0.822) (40,0.828) (45,0.83) (50,0.831) (55,0.832) (60,0.833) (65,0.833) (70,0.851) (75,0.837) (80,0.83) (85,0.857) (90,0.85) (95,0.86) (100,0.86)
};
\addplot coordinates {
(0,0.49) (1,0.519) (2,0.518) (3,0.519) (4,0.53) (5,0.533) (6,0.543) (7,0.547) (8,0.538) (9,0.556) (10,0.55) (15,0.633) (20,0.63) (25,0.676) (30,0.783) (35,0.89) (40,0.919) (45,0.947) (50,0.95) (55,0.972) (60,0.97) (65,0.978) (70,0.983) (75,0.977) (80,0.986) (85,0.98) (90,0.988) (95,0.987) (100,0.98)
};
\addplot coordinates {
(0,0.498) (1,0.521) (2,0.527) (3,0.532) (4,0.55) (5,0.551) (6,0.569) (7,0.569) (8,0.566) (9,0.586) (10,0.55) (15,0.637) (20,0.654) (25,0.684) (30,0.756) (35,0.845) (40,0.8) (45,0.877) (50,0.896) (55,0.922) (60,0.938) (65,0.945) (70,0.953) (75,0.956) (80,0.959) (85,0.955) (90,0.96) (95,0.96) (100,0.968)
};
\pgfplotsset{cycle list shift=1}
\addplot coordinates {
(0,0.49) (1,0.501) (2,0.506) (3,0.0) (4,0.498) (5,0.505) (6,0.5) (7,0.516) (8,0.503) (9,0.523) (10,0.513) (15,0.569) (20,0.614) (25,0.678) (30,0.724) (35,0.751) (40,0.828) (45,0.872) (50,0.93) (55,0.934) (60,0.939) (65,0.941) (70,0.942) (75,0.942) (80,0.946) (85,0.944) (90,0.939) (95,0.941) (100,0.949)
};
\addplot+[color=black!90,dashed] coordinates {
(0,0.49731999999999993) (1,0.52782) (2,0.53764) (3,0.53408) (4,0.56474) (5,0.5656199999999999) (6,0.56334) (7,0.56998) (8,0.57234) (9,0.58792) (10,0.58102) (15,0.64028) (20,0.6671400000000001) (25,0.73738) (30,0.8084) (35,0.8620599999999999) (40,0.88612) (45,0.9056000000000001) (50,0.92318) (55,0.93204) (60,0.9363199999999999) (65,0.9398199999999999) (70,0.94612) (75,0.94154) (80,0.94628) (85,0.9487) (90,0.9498999999999999) (95,0.95106) (100,0.95156)
};
\legend{existential,logical,numbers,quantifiers,implicit-rel,overall}
\end{axis}
\end{tikzpicture}}
\end{center}

\begin{center}
\resizebox{0.96\linewidth}{!}{
\begin{tikzpicture}
\begin{axis}[
every axis plot/.append style={line width=2.5pt},
cycle list name=Dark2,
mark options={mark=none},
width=10cm,
height=7.5cm,
xmin=0, xmax=100,
xtick={0,20,40,60,80,100},
xticklabel style={font=\footnotesize},
xmajorgrids,
ymin=0.49, ymax=1.0,
ytick={0.5,0.6,0.7,0.8,0.9,1.0},
ymajorgrids,
yticklabel style={font=\footnotesize},
grid style=dashed,
legend style={at={(1.0,0.5)},font=\scriptsize}
]
\addplot coordinates {
(0,0.508) (1,0.524) (2,0.558) (3,0.56) (4,0.612) (5,0.596) (6,0.597) (7,0.611) (8,0.62) (9,0.626) (10,0.632) (15,0.68) (20,0.705) (25,0.739) (30,0.761) (35,0.874) (40,0.965) (45,0.985) (50,0.99) (55,0.997) (60,0.998) (65,0.999) (70,0.997) (75,0.999) (80,0.999) (85,0.997) (90,0.999) (95,0.999) (100,0.999)
};
\addplot coordinates {
(0,0.499) (1,0.519) (2,0.548) (3,0.526) (4,0.569) (5,0.541) (6,0.552) (7,0.605) (8,0.614) (9,0.634) (10,0.6) (15,0.636) (20,0.636) (25,0.652) (30,0.662) (35,0.719) (40,0.79) (45,0.829) (50,0.843) (55,0.849) (60,0.852) (65,0.854) (70,0.853) (75,0.85) (80,0.854) (85,0.853) (90,0.85) (95,0.855) (100,0.85)
};
\addplot coordinates {
(0,0.51) (1,0.507) (2,0.532) (3,0.552) (4,0.555) (5,0.556) (6,0.539) (7,0.558) (8,0.559) (9,0.551) (10,0.557) (15,0.615) (20,0.63) (25,0.644) (30,0.671) (35,0.703) (40,0.758) (45,0.822) (50,0.891) (55,0.926) (60,0.921) (65,0.943) (70,0.956) (75,0.965) (80,0.95) (85,0.966) (90,0.973) (95,0.981) (100,0.983)
};
\addplot coordinates {
(0,0.489) (1,0.53) (2,0.53) (3,0.57) (4,0.572) (5,0.568) (6,0.566) (7,0.58) (8,0.589) (9,0.588) (10,0.59) (15,0.62) (20,0.643) (25,0.644) (30,0.668) (35,0.704) (40,0.771) (45,0.823) (50,0.872) (55,0.911) (60,0.92) (65,0.939) (70,0.941) (75,0.955) (80,0.947) (85,0.958) (90,0.966) (95,0.967) (100,0.963)
};
\pgfplotsset{cycle list shift=2}
\addplot coordinates {
(0,0.512) (1,0.495) (2,0.503) (3,0.499) (4,0.499) (5,0.497) (6,0.504) (7,0.493) (8,0.505) (9,0.502) (10,0.523) (15,0.567) (20,0.5) (25,0.614) (30,0.692) (35,0.78) (40,0.871) (45,0.902) (50,0.926) (55,0.94) (60,0.938) (65,0.941) (70,0.943) (75,0.947) (80,0.944) (85,0.945) (90,0.95) (95,0.951) (100,0.952)
};
\addplot+[color=black!90,dashed] coordinates {
(0,0.50398) (1,0.51598) (2,0.5361400000000001) (3,0.5431) (4,0.562) (5,0.5521199999999999) (6,0.552) (7,0.56986) (8,0.5788) (9,0.5806799999999999) (10,0.5808199999999999) (15,0.6266200000000001) (20,0.6372599999999999) (25,0.6591199999999999) (30,0.69148) (35,0.75692) (40,0.83206) (45,0.8725400000000001) (50,0.9056599999999999) (55,0.9252) (60,0.92642) (65,0.9357199999999999) (70,0.9386399999999998) (75,0.9446200000000001) (80,0.93954) (85,0.94436) (90,0.94892) (95,0.95106) (100,0.9510200000000001)
};
\legend{existential,logical,numbers,quantifiers,superlatives,overall}
\end{axis}
\end{tikzpicture}}
\end{center}

\begin{center}
\resizebox{0.96\linewidth}{!}{
\begin{tikzpicture}
\begin{axis}[
every axis plot/.append style={line width=2.5pt},
cycle list name=Dark2,
mark options={mark=none},
width=10cm,
height=7.5cm,
xmin=0, xmax=100,
xtick={0,20,40,60,80,100},
xticklabel style={font=\footnotesize},
xmajorgrids,
ymin=0.49, ymax=1.0,
ytick={0.5,0.6,0.7,0.8,0.9,1.0},
ymajorgrids,
yticklabel style={font=\footnotesize},
grid style=dashed,
legend style={at={(1.0,1.0)},font=\scriptsize}
]
\addplot coordinates {
(0,0.501) (1,0.5) (2,0.518) (3,0.537) (4,0.552) (5,0.597) (6,0.574) (7,0.54) (8,0.542) (9,0.554) (10,0.552) (15,0.528) (20,0.499) (25,0.526) (30,0.541) (35,0.5) (40,0.519) (45,0.516) (50,0.526) (55,0.54) (60,0.6) (65,0.616) (70,0.626) (75,0.627) (80,0.64) (85,0.68) (90,0.705) (95,0.731) (100,0.76)
};
\addplot coordinates {
(0,0.5) (1,0.511) (2,0.513) (3,0.52) (4,0.526) (5,0.547) (6,0.521) (7,0.505) (8,0.517) (9,0.518) (10,0.519) (15,0.5) (20,0.506) (25,0.517) (30,0.515) (35,0.512) (40,0.507) (45,0.512) (50,0.5) (55,0.51) (60,0.573) (65,0.603) (70,0.629) (75,0.635) (80,0.64) (85,0.642) (90,0.642) (95,0.65) (100,0.645)
};
\addplot coordinates {
(0,0.509) (1,0.508) (2,0.527) (3,0.517) (4,0.51) (5,0.509) (6,0.52) (7,0.505) (8,0.512) (9,0.496) (10,0.503) (15,0.508) (20,0.508) (25,0.505) (30,0.513) (35,0.495) (40,0.508) (45,0.49) (50,0.501) (55,0.527) (60,0.515) (65,0.539) (70,0.56) (75,0.56) (80,0.565) (85,0.59) (90,0.609) (95,0.628) (100,0.635)
};
\addplot coordinates {
(0,0.501) (1,0.507) (2,0.526) (3,0.532) (4,0.545) (5,0.522) (6,0.531) (7,0.533) (8,0.517) (9,0.528) (10,0.525) (15,0.523) (20,0.496) (25,0.508) (30,0.517) (35,0.5) (40,0.5) (45,0.504) (50,0.5) (55,0.531) (60,0.539) (65,0.576) (70,0.598) (75,0.594) (80,0.6) (85,0.628) (90,0.63) (95,0.64) (100,0.649)
};
\addplot coordinates {
(0,0.5) (1,0.514) (2,0.495) (3,0.494) (4,0.506) (5,0.503) (6,0.499) (7,0.5) (8,0.499) (9,0.497) (10,0.503) (15,0.495) (20,0.507) (25,0.48) (30,0.496) (35,0.496) (40,0.49) (45,0.495) (50,0.505) (55,0.503) (60,0.49) (65,0.493) (70,0.493) (75,0.504) (80,0.521) (85,0.564) (90,0.575) (95,0.6) (100,0.62)
};
\addplot coordinates {
(0,0.5) (1,0.51) (2,0.503) (3,0.504) (4,0.499) (5,0.49) (6,0.496) (7,0.49) (8,0.5) (9,0.502) (10,0.506) (15,0.49) (20,0.503) (25,0.505) (30,0.491) (35,0.497) (40,0.497) (45,0.499) (50,0.492) (55,0.499) (60,0.49) (65,0.526) (70,0.515) (75,0.52) (80,0.533) (85,0.545) (90,0.576) (95,0.58) (100,0.607)
};
\addplot+[color=black!90,dashed] coordinates {
(0,0.5032333333333334) (1,0.5178666666666667) (2,0.5141166666666667) (3,0.5179999999999999) (4,0.5238) (5,0.5298166666666667) (6,0.52405) (7,0.51425) (8,0.5151333333333333) (9,0.51625) (10,0.51855) (15,0.5104833333333334) (20,0.5035333333333333) (25,0.5087166666666667) (30,0.5127) (35,0.5067) (40,0.5064000000000001) (45,0.50435) (50,0.5060833333333333) (55,0.5192500000000001) (60,0.5374666666666666) (65,0.5594666666666667) (70,0.5707333333333334) (75,0.57515) (80,0.5847666666666667) (85,0.6101500000000001) (90,0.6232833333333333) (95,0.63945) (100,0.6531666666666666)
};
\legend{existential,logical,numbers,quantifiers,relational,implicit-rel,overall}
\end{axis}
\end{tikzpicture}}
\end{center}

\subsection{Distribution simple-spatial vs relational}\label{appendix:distribution}

\begin{center}
\resizebox{0.96\linewidth}{!}{
\begin{tikzpicture}
\begin{axis}[
every axis plot/.append style={line width=2.5pt},
cycle list name=Dark2,
mark options={mark=none},
width=10cm,
height=7.5cm,
xmin=0, xmax=100,
xtick={0,20,40,60,80,100},
xticklabel style={font=\footnotesize},
xmajorgrids,
ymin=0.49, ymax=1.0,
ytick={0.5,0.6,0.7,0.8,0.9,1.0},
ymajorgrids,
yticklabel style={font=\footnotesize},
grid style=dashed,
legend style={at={(1.0,0.6)},font=\scriptsize}
]
\addplot coordinates {
(0,0.49675) (1,0.5014000000000001) (2,0.50165) (3,0.496) (4,0.5009) (5,0.4959) (6,0.5029) (7,0.5056499999999999) (8,0.5085999999999999) (9,0.5032) (10,0.5296000000000001) (15,0.50945) (20,0.5095000000000001) (25,0.5002) (30,0.5048) (35,0.49465) (40,0.5002) (45,0.51035) (50,0.49895) (55,0.4995) (60,0.5192) (65,0.5284) (70,0.50645) (75,0.5058999999999999) (80,0.5223) (85,0.53765) (90,0.5141000000000001) (95,0.52305) (100,0.54985)
};
\addplot coordinates {
(0,0.5002) (1,0.5127999999999999) (2,0.5102500000000001) (3,0.49970000000000003) (4,0.52565) (5,0.5108999999999999) (6,0.511) (7,0.5178) (8,0.5245) (9,0.5201) (10,0.5327) (15,0.51525) (20,0.5266500000000001) (25,0.53305) (30,0.5285500000000001) (35,0.51795) (40,0.5016) (45,0.53345) (50,0.52885) (55,0.5262) (60,0.50905) (65,0.51835) (70,0.5267999999999999) (75,0.5063500000000001) (80,0.53115) (85,0.51185) (90,0.5116499999999999) (95,0.5324) (100,0.5306)
};
\addplot coordinates {
(0,0.49) (1,0.51) (2,0.522) (3,0.515) (4,0.504) (5,0.521) (6,0.52) (7,0.531) (8,0.516) (9,0.534) (10,0.535) (15,0.584) (20,0.0) (25,0.625) (30,0.634) (35,0.64) (40,0.63) (45,0.658) (50,0.667) (55,0.696) (60,0.728) (65,0.745) (70,0.782) (75,0.82) (80,0.857) (85,0.885) (90,0.91) (95,0.919) (100,0.935)
};
\addplot coordinates {
(0,0.4976) (1,0.5104) (2,0.5154000000000001) (3,0.51505) (4,0.5152000000000001) (5,0.52225) (6,0.51135) (7,0.5296000000000001) (8,0.5247999999999999) (9,0.52435) (10,0.54435) (15,0.5862499999999999) (20,0.5942) (25,0.6303) (30,0.64115) (35,0.65325) (40,0.65375) (45,0.6636) (50,0.67035) (55,0.66185) (60,0.6718500000000001) (65,0.6875) (70,0.6959) (75,0.6998) (80,0.7098) (85,0.7327999999999999) (90,0.74505) (95,0.76715) (100,0.78435)
};
\addplot coordinates {
(0,0.5023) (1,0.5103500000000001) (2,0.5397) (3,0.5578000000000001) (4,0.5742499999999999) (5,0.5939000000000001) (6,0.604) (7,0.6033999999999999) (8,0.6057999999999999) (9,0.6067) (10,0.61155) (15,0.63175) (20,0.64495) (25,0.6613500000000001) (30,0.661) (35,0.67275) (40,0.6871499999999999) (45,0.69645) (50,0.705) (55,0.7286) (60,0.7597499999999999) (65,0.7778) (70,0.7976) (75,0.8110999999999999) (80,0.8188500000000001) (85,0.8323499999999999) (90,0.8388500000000001) (95,0.845) (100,0.85835)
};
\addplot coordinates {
(0,0.49985) (1,0.5099) (2,0.52885) (3,0.5325) (4,0.53225) (5,0.5562) (6,0.5580499999999999) (7,0.5555) (8,0.55725) (9,0.56975) (10,0.5446) (15,0.56785) (20,0.61425) (25,0.6371) (30,0.6511) (35,0.6661) (40,0.67685) (45,0.67655) (50,0.67955) (55,0.6852499999999999) (60,0.69355) (65,0.6978) (70,0.70555) (75,0.7172499999999999) (80,0.7200500000000001) (85,0.7478) (90,0.76675) (95,0.7823) (100,0.7903500000000001)
};
\addplot coordinates {
(0,0.4998) (1,0.5192) (2,0.5281499999999999) (3,0.5177) (4,0.52005) (5,0.52635) (6,0.53905) (7,0.51465) (8,0.5144) (9,0.50125) (10,0.5249) (15,0.5146) (20,0.52285) (25,0.5088) (30,0.52755) (35,0.51215) (40,0.5015000000000001) (45,0.5197) (50,0.54085) (55,0.5206500000000001) (60,0.5444) (65,0.5488) (70,0.51235) (75,0.5312) (80,0.55585) (85,0.54765) (90,0.5696000000000001) (95,0.5427) (100,0.54005)
};
\legend{45\%,47.5\%,50\%,52.5\%,55\%,57.5\%,60\%}
\end{axis}
\end{tikzpicture}}
\end{center}

\newpage

\section*{Acknowledgments}

We thank the anonymous reviewers for their constructive feedback. AK is grateful for being supported by a Qualcomm Research Studentship and an EPSRC Doctoral Training Studentship.

\bibliographystyle{acl_natbib_nourl}
\bibliography{bibliography}

\begin{thebibliography}{35}
\expandafter\ifx\csname natexlab\endcsname\relax\def\natexlab#1{#1}\fi

\bibitem[{Agrawal et~al.(2016)Agrawal, Batra, and Parikh}]{Agrawal2016}
Aishwarya Agrawal, Dhruv Batra, and Devi Parikh. 2016.
\newblock Analyzing the behavior of visual question answering models.
\newblock In \emph{Proceedings of the 2016 Conference on Empirical Methods in
  Natural Language Processing}, EMNLP 2016, pages 1955--1960.

\bibitem[{Andreas et~al.(2016)Andreas, Rohrbach, Darrell, and
  Klein}]{Andreas2016a}
Jacob Andreas, Marcus Rohrbach, Trevor Darrell, and Dan Klein. 2016.
\newblock Neural module networks.
\newblock In \emph{Proceedings of the IEEE Conference on Computer Vision and
  Pattern Recognition}, CVPR 2016.

\bibitem[{Antol et~al.(2015)Antol, Agrawal, Lu, Mitchell, Batra, Zitnick, and
  Parikh}]{Antol2015}
Stanislaw Antol, Aishwarya Agrawal, Jiasen Lu, Margaret Mitchell, Dhruv Batra,
  C.~Lawrence Zitnick, and Devi Parikh. 2015.
\newblock {VQA}: {V}isual question answering.
\newblock In \emph{Proceedings of the IEEE International Conference on Computer
  Vision}, ICCV 2015.

\bibitem[{Bengio et~al.(2009)Bengio, Louradour, Collobert, and
  Weston}]{Bengio2009}
Yoshua Bengio, J{\'e}r\^{o}me Louradour, Ronan Collobert, and Jason Weston.
  2009.
\newblock Curriculum learning.
\newblock In \emph{Proceedings of the 26th Annual International Conference on
  Machine Learning}, ICML 2009, pages 41--48.

\bibitem[{Bowman et~al.(2015)Bowman, Angeli, Potts, and Manning}]{Bowman2015b}
Samuel~R. Bowman, Gabor Angeli, Christopher Potts, and Christopher~D. Manning.
  2015.
\newblock A large annotated corpus for learning natural language inference.
\newblock In \emph{Proceedings of the Conference on Empirical Methods in
  Natural Language Processing}, EMNLP 2015, pages 632--642.

\bibitem[{Elman(1993)}]{Elman1993}
Jeffrey~L. Elman. 1993.
\newblock Learning and development in neural networks: the importance of
  starting small.
\newblock \emph{Cognition}, 48(1):71--99.

\bibitem[{Gers and Schmidhuber(2001)}]{Gers2001}
Felix~A. Gers and J\"{u}rgen Schmidhuber. 2001.
\newblock {LSTM} recurrent networks learn simple context-free and
  context-sensitive languages.
\newblock \emph{Transactions on Neural Networks}, 12(6):1333--1340.

\bibitem[{Goyal et~al.(2017)Goyal, Khot, Summers-Stay, Batra, and
  Parikh}]{Goyal2017}
Yash Goyal, Tejas Khot, Douglas Summers-Stay, Dhruv Batra, and Devi Parikh.
  2017.
\newblock Making the {V} in {VQA} matter: {E}levating the role of image
  understanding in {V}isual {Q}uestion {A}nswering.
\newblock In \emph{Proceedings of the IEEE Conference on Computer Vision and
  Pattern Recognition}, CVPR 2017, pages 6325--6334.

\bibitem[{Halevy et~al.(2009)Halevy, Norvig, and Pereira}]{Halevy2009}
Alon Halevy, Peter Norvig, and Fernando Pereira. 2009.
\newblock The unreasonable effectiveness of data.
\newblock \emph{IEEE Intelligent Systems}, 24(2):8--12.

\bibitem[{Henderson et~al.(2018)Henderson, Islam, Bachman, Pineau, Precup, and
  Meger}]{Henderson2018}
Peter Henderson, Riashat Islam, Philip Bachman, Joelle Pineau, Doina Precup,
  and David Meger. 2018.
\newblock Deep reinforcement learning that matters.
\newblock In \emph{{AAAI}}. {AAAI} Press.

\bibitem[{Hodosh and Hockenmaier(2016)}]{Hodosh2016}
Micah Hodosh and Julia Hockenmaier. 2016.
\newblock Focused evaluation for image description with binary forced-choice
  tasks.
\newblock In \emph{Proceedings of the 5th Workshop on Vision and Language},
  pages 19--28.

\bibitem[{Hu et~al.(2017)Hu, Andreas, Rohrbach, Darrell, and Saenko}]{Hu2017}
Ronghang Hu, Jacob Andreas, Marcus Rohrbach, Trevor Darrell, and Kate Saenko.
  2017.
\newblock Learning to reason: {E}nd-to-end module networks for visual question
  answering.
\newblock In \emph{Proceedings of the IEEE International Conference on Computer
  Vision}, ICCV 2017.

\bibitem[{Hudson and Manning(2018)}]{Hudson2018}
Drew~A. Hudson and Christopher~D. Manning. 2018.
\newblock Compositional attention networks for machine reasoning.
\newblock In \emph{Proceedings of the International Conference on Learning
  Representations}, ICLR 2018.

\bibitem[{Jia and Liang(2017)}]{Jia2017}
Robin Jia and Percy Liang. 2017.
\newblock Adversarial examples for evaluating reading comprehension systems.
\newblock In \emph{Proceedings of the Conference on Empirical Methods in
  Natural Language Processing}, EMNLP 2017, pages 2021--2031.

\bibitem[{Johnson et~al.(2017{\natexlab{a}})Johnson, Hariharan, van~der Maaten,
  Fei-Fei, Zitnick, and Girshick}]{Johnson2017a}
Justin Johnson, Bharath Hariharan, Laurens van~der Maaten, Li~Fei-Fei,
  C.~Lawrence Zitnick, and Ross Girshick. 2017{\natexlab{a}}.
\newblock {CLEVR}: {A} diagnostic dataset for compositional language and
  elementary visual reasoning.
\newblock In \emph{Proceedings of the IEEE Conference on Computer Vision and
  Pattern Recognition}, CVPR 2017.

\bibitem[{Johnson et~al.(2017{\natexlab{b}})Johnson, Hariharan, van~der Maaten,
  Hoffman, Fei-Fei, Zitnick, and Girshick}]{Johnson2017b}
Justin Johnson, Bharath Hariharan, Laurens van~der Maaten, Judy Hoffman,
  Li~Fei-Fei, C~Lawrence Zitnick, and Ross Girshick. 2017{\natexlab{b}}.
\newblock Inferring and executing programs for visual reasoning.
\newblock In \emph{Proceedings of the IEEE International Conference on Computer
  Vision}, ICCV 2017.

\bibitem[{Joulin and Mikolov(2015)}]{Joulin2015}
Armand Joulin and Tomas Mikolov. 2015.
\newblock Inferring algorithmic patterns with stack-augmented recurrent nets.
\newblock In \emph{Advances in Neural Information Processing Systems 28}, pages
  190--198. Curran Associates, Inc.

\bibitem[{Kuhnle and Copestake(2017)}]{Kuhnle2017}
Alexander Kuhnle and Ann Copestake. 2017.
\newblock Shape{W}orld - {A} new test methodology for multimodal language
  understanding.
\newblock \emph{ArXiv e-prints 1704.04517}.

\bibitem[{Kuhnle and Copestake(2018)}]{Kuhnle2018}
Alexander Kuhnle and Ann Copestake. 2018.
\newblock Deep learning evaluation using deep linguistic processing.
\newblock In \emph{Proceedings of the Workshop on Generalization in the Age of
  Deep Learning}, NAACL 2018, pages 17--23.

\bibitem[{Lake and Baroni(2017)}]{Lake2017}
Brenden~M. Lake and Marco Baroni. 2017.
\newblock Still not systematic after all these years: {O}n the compositional
  skills of sequence-to-sequence recurrent networks.
\newblock \emph{ArXiv e-prints 1711.00350}.

\bibitem[{Lucic et~al.(2017)Lucic, Kurach, Michalski, Gelly, and
  Bousquet}]{Lucic2017}
Mario Lucic, Karol Kurach, Marcin Michalski, Sylvain Gelly, and Olivier
  Bousquet. 2017.
\newblock Are {GAN}s created equal? {A} large-scale study.
\newblock \emph{CoRR}, abs/1711.10337.

\bibitem[{Mascharka et~al.(2018)Mascharka, Tran, Soklaski, and
  Majumdar}]{Mascharka2018}
David Mascharka, Philip Tran, Ryan Soklaski, and Arjun Majumdar. 2018.
\newblock Transparency by design: {C}losing the gap between performance and
  interpretability in visual reasoning.
\newblock In \emph{Proceedings of the IEEE Conference on Computer Vision and
  Pattern Recognition}, CVPR 2018.

\bibitem[{Melis et~al.(2018)Melis, Dyer, and Blunsom}]{Melis2018}
G\'{a}bor Melis, Chris Dyer, and Phil Blunsom. 2018.
\newblock On the state of the art of evaluation in neural language models.
\newblock In \emph{Proceedings of the International Conference on Learning
  Representations}, ICLR 2018.

\bibitem[{Merity et~al.(2018)Merity, Keskar, and Socher}]{Merity2018}
Stephen Merity, Nitish~Shirish Keskar, and Richard Socher. 2018.
\newblock An analysis of neural language modeling at multiple scales.
\newblock \emph{ArXiv e-prints 1803.08240}.

\bibitem[{Mudrakarta et~al.(2018)Mudrakarta, Taly, Sundararajan, and
  Dhamdhere}]{Mudrakarta2018}
Pramod~Kaushik Mudrakarta, Ankur Taly, Mukund Sundararajan, and Kedar
  Dhamdhere. 2018.
\newblock Did the model understand the question?
\newblock \emph{ArXiv e-prints 1805.05492}.

\bibitem[{Perez et~al.(2018)Perez, Strub, de~Vries, Dumoulin, and
  Courville}]{Perez2018}
Ethan Perez, Florian Strub, Harm de~Vries, Vincent Dumoulin, and Aaron~C.
  Courville. 2018.
\newblock {FiLM}: {V}isual reasoning with a general conditioning layer.
\newblock In \emph{AAAI}.

\bibitem[{Poliak et~al.(2018)Poliak, Naradowsky, Haldar, Rudinger, and {Van
  Durme}}]{Poliak2018}
Adam Poliak, Jason Naradowsky, Aparajita Haldar, Rachel Rudinger, and Benjamin
  {Van Durme}. 2018.
\newblock Hypothesis only baselines for natural language inference.
\newblock In \emph{The Seventh Joint Conference on Lexical and Computational
  Semantics}, *SEM 2018.

\bibitem[{Rajpurkar et~al.(2016)Rajpurkar, Zhang, Lopyrev, and
  Liang}]{Rajpurkar2016}
Pranav Rajpurkar, Jian Zhang, Konstantin Lopyrev, and Percy Liang. 2016.
\newblock {SQuAD}: 100,000+ questions for machine comprehension of text.
\newblock In \emph{Proceedings of the 2016 Conference on Empirical Methods in
  Natural Language Processing}, EMNLP 2016, pages 2383--2392.

\bibitem[{Santoro et~al.(2017)Santoro, Raposo, Barrett, Malinowski, Pascanu,
  Battaglia, and Lillicrap}]{Santoro2017}
Adam Santoro, David Raposo, David G.~T. Barrett, Mateusz Malinowski, Razvan
  Pascanu, Peter Battaglia, and Timothy~P. Lillicrap. 2017.
\newblock A simple neural network module for relational reasoning.
\newblock In \emph{Advances in Neural Information Processing Systems 30: Annual
  Conference on Neural Information Processing Systems 2017}, pages 4974--4983.

\bibitem[{Shekhar et~al.(2017)Shekhar, Pezzelle, Klimovich, Herbelot, Nabi,
  Sangineto, and Bernardi}]{Shekhar2017}
Ravi Shekhar, Sandro Pezzelle, Yauhen Klimovich, Aur\'{e}lie Herbelot, Moin
  Nabi, Enver Sangineto, and Raffaella Bernardi. 2017.
\newblock {FOIL} it! {F}ind one mismatch between image and language caption.
\newblock In \emph{Proceedings of the 55th Annual Meeting of the Association
  for Computational Linguistics}, ACL 2017, pages 255--265.

\bibitem[{Suarez et~al.(2018)Suarez, Johnson, and Li}]{Suarez2018}
Joseph Suarez, Justin Johnson, and Fei{-}Fei Li. 2018.
\newblock {DDRprog}: {A} {CLEVR} differentiable dynamic reasoning programmer.
\newblock \emph{ArXiv e-prints 1803.11361}.

\bibitem[{Suhr et~al.(2017)Suhr, Lewis, Yeh, and Artzi}]{Suhr2017}
Alane Suhr, Mike Lewis, James Yeh, and Yoav Artzi. 2017.
\newblock A corpus of natural language for visual reasoning.
\newblock In \emph{55th Annual Meeting of the Association for Computational
  Linguistics}, ACL 2017.

\bibitem[{Weston et~al.(2015)Weston, Bordes, Chopra, and Mikolov}]{Weston2015}
Jason Weston, Antoine Bordes, Sumit Chopra, and Tomas Mikolov. 2015.
\newblock Towards {AI}-complete question answering: {A} set of prerequisite toy
  tasks.
\newblock \emph{ArXiv e-prints 1502.05698}.

\bibitem[{Yang et~al.(2018)Yang, Ganichev, Wang, Shlens, and
  Sussillo}]{Yang2018}
Guangyu~Robert Yang, Igor Ganichev, Xiao-Jing Wang, Jonathon Shlens, and David
  Sussillo. 2018.
\newblock A dataset and architecture for visual reasoning with a working
  memory.
\newblock \emph{ArXiv e-prints 1803.06092}.

\bibitem[{Yang et~al.(2016)Yang, He, Gao, Deng, and Smola}]{Yang2016}
Zichao Yang, Xiaodong He, Jianfeng Gao, Li~Deng, and Alexander~J. Smola. 2016.
\newblock Stacked attention networks for image question answering.
\newblock In \emph{Proceedings of the IEEE Conference on Computer Vision and
  Pattern Recognition}, CVPR 2016.

\end{thebibliography}

\end{document}